\documentclass[lettersize,journal]{IEEEtran}
\usepackage{amsmath,amsfonts}
\usepackage{algorithmic}
\usepackage{algorithm}
\usepackage{array}
\usepackage[caption=false,font=normalsize,labelfont=sf,textfont=sf]{subfig}
\usepackage{textcomp}
\usepackage{stfloats}
\usepackage{url}
\usepackage{verbatim}
\usepackage{graphicx}
\usepackage{cite}
\usepackage{color}
\usepackage{ulem}
\usepackage{here}
\usepackage{empheq}

\begin{document}

\title{Probabilistic Semantic Data Association \\ for Collaborative Human-Robot Sensing}

\author{Shohei Wakayama and Nisar Ahmed$^*$
\thanks{$^*$The authors are with the Smead Aerospace Engineering Sciences Department, University of Colorado Boulder, Boulder, CO 80303 USA        {\tt\small [shohei.wakayama; nisar.ahmed]@colorado.edu}}
}

\markboth{IEEE Trans. Robotics,~Vol.~V, No.~N, December~2022}%
{Wakayama and Ahmed: Probabilistic Semantic Data Association}


\maketitle

\begin{abstract}
\textcolor{black}
{ 
Humans cannot always be treated as oracles for collaborative sensing. Robots thus need to maintain beliefs over unknown world states when receiving semantic data from humans, as well as account for possible discrepancies between human-provided data and these beliefs.} 
To this end, this paper {introduces the problem} of semantic data association (SDA) in relation to conventional data association problems for sensor fusion. It then develops a novel probabilistic semantic data association (PSDA) algorithm to rigorously address SDA in general settings, unlike previous work on semantic data fusion which developed heuristic techniques for specific settings. PSDA is further incorporated into a recursive hybrid Bayesian data fusion scheme which uses Gaussian mixture priors for object states and softmax functions for semantic human sensor data likelihoods. Simulations of a  multi-object search task 
show that PSDA enables robust collaborative state estimation under a wide range of conditions where semantic human sensor data can be erroneous or contain significant reference ambiguities. 
\end{abstract}

\begin{IEEEkeywords}
Data association, semantic data fusion, human-robot interaction, multi-object search, state estimation.
\end{IEEEkeywords}

\section{Introduction}
{\IEEEPARstart{M}{obile} robots 
are expected to operate autonomously in hazardous and} uncertain environments, such as other planetary surfaces and disaster areas \cite{gao2019, tsitsimpelis2019, bailon-ruiz2020}. With limits on sensing, computation, and actuation from size, weight, power, and cost constraints, humans must \textcolor{black}{supervise robots} in these settings. This can result in heavy mental burdens, leading to inefficiencies and degraded task performance. 
{Research suggests that effectiveness can be improved by allowing humans to act as information providers for autonomous robots, either instead of or in addition to \textcolor{black}{commanding them} \cite{lewis2009}}. 
Other works have also examined using `human sensor' data to augment robotic state estimation in uncertain environments \cite{kaupp2005_adaptive, kaupp2005_operator, kaupp2007, rosenthal2011}, and in particular how Bayesian fusion of {\it soft semantic data} provided by humans (e.g. `the car is behind the wall') with {\it hard} sensor data 
(radar, lidar, vision, etc.) can  improve 
object localization 
\cite{nisarTRO, lukeTRO, lukeRAS, lukeFUSION}.

\begin{figure}[t]
    \centering
    \includegraphics[width=6.5cm]{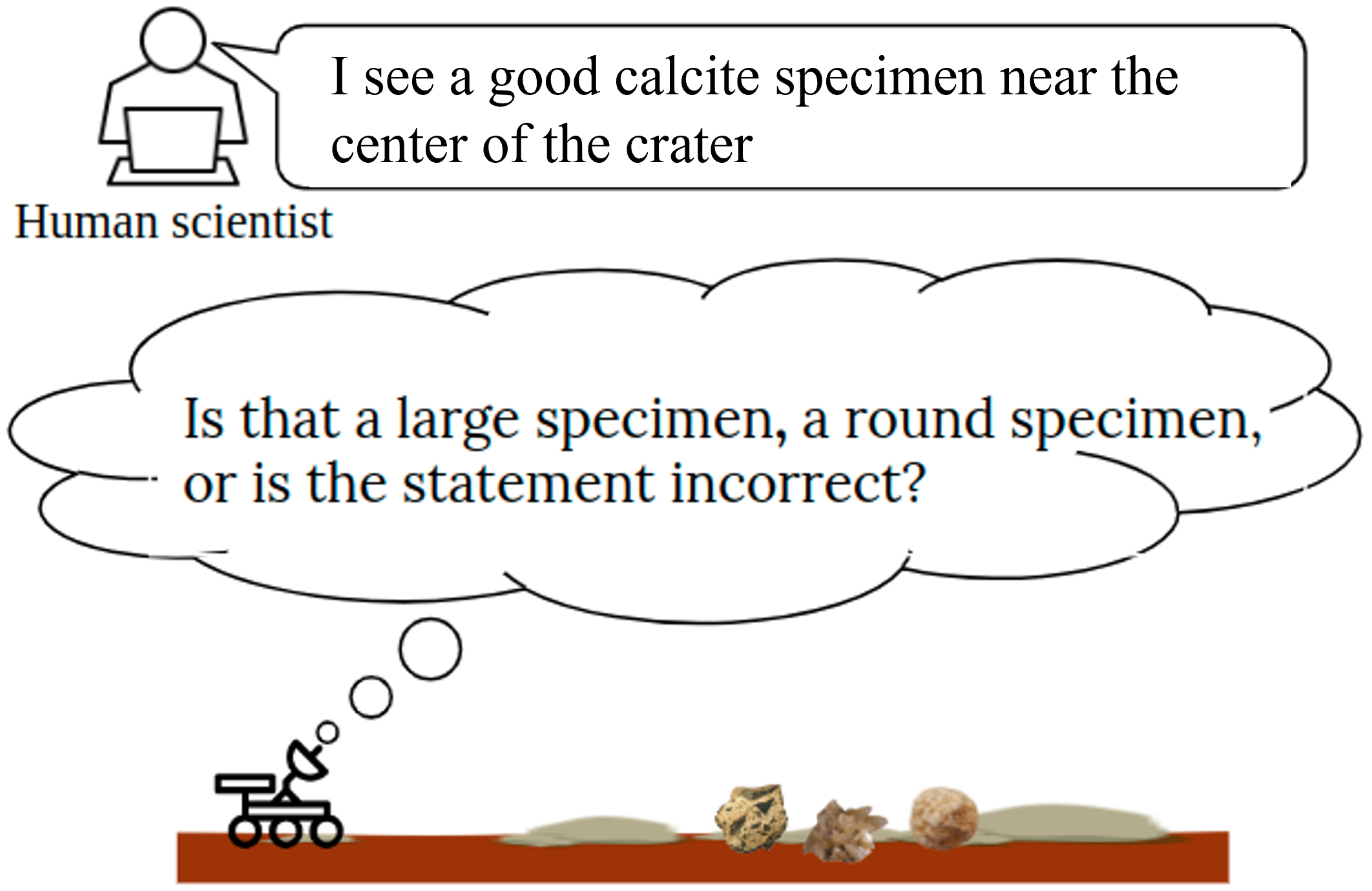}
    \caption{Example planetary science survey scenario where \textcolor{magenta}{a} human collaborator provides soft semantic data about interesting mineral targets: to localize specimens with desirable mineral types and size/shape, the robot must interpret possibly erroneous or ambiguous data for targets which may or may not exist.}
    \label{fig:data_association_example}
    \vspace{-0.25in}
\end{figure}


Such approaches to collaborative sensing  
raise other challenges. Some of these arise due to the fact that in many cases humans must also perform other tasks in parallel with sensing, leading to a strain on cognitive resources that also opens the door to mistakes and biases \cite{ricks2004}. Humans may also encounter difficulties \textcolor{black}{in} interpreting resources for gathering soft data, such as images in a robot's sensor field of view (FOV). 
For instance, consider the situation in Fig. \ref{fig:data_association_example}, in which a remotely located human scientist and an autonomous mobile robot collaborate on a futuristic planetary survey mission to search for specific types of scientifically interesting rock specimens. This scenario is based on the idea of deep space telepresence \cite{lester2017telepresence}, where the human provides semantic data about target specimen location and type properties, which can be fused together with other sensor data to narrow down survey areas. Here, the human could mistake an uninteresting object in the robot's camera FOV for an interesting one, or fail to fully convey specimen features, leading to ambiguity about the kind of rock being referenced. If the robot naively fuses this soft data into specimen state estimates, any resulting errors could make it harder to find the desired rocks and degrade performance. While the human could always be prompted to double-check their inputs, 
this would create added delays 
and workload. 
Thus, without human oracles, 
robots must account for erroneous or incomplete soft semantic data during fusion. 


{\it Data association (DA)} methods consider uncertainties in the origins of observations and have
been widely explored for target tracking \cite{bar-shalom1975, bar-shalom2009, vo2015}. However, conventional DA algorithms handle only continuous hard sensor data as opposed to semantic data about continuous 
states \textcolor{black}{
that correspond to non-Gaussian state uncertainties, which violate typical DA algorithm assumptions} 
\cite{redmon2016}. Also, existing works on managing uncertain semantic data \cite{bishop2011, mahler2007}  
cannot reason about reference object associations in a 
manner that meshes well with 
existing robotic soft-hard data fusion methods. 
Against this backdrop, this paper's main contributions are: 1) formal definition of {\it semantic data association (SDA)} as the class of state estimation problems that must cope with potentially erroneous/incomplete semantic data; and 2) development of the {\it probabilistic semantic data association (PSDA)} algorithm to address SDA, via novel theoretical extension to the well-known probabilistic data association method used for continuous hard data fusion \cite{bar-shalom2009, schulz2001}.
%
%
%

PSDA provides a rigorous solution to SDA by inferring dynamic association hypothesis probabilities for semantic data in Bayesian fusion updates of estimated object states. These dynamic probabilities avoid incorrectly biasing or collapsing object state posterior distributions in the presence of imperfect semantic sensing, thereby improving resulting estimates. 
A key advantage to tackling SDA in general -- and computing dynamic association hypothesis weights in particular -- is that PSDA can efficiently exploit information generated by existing 
soft-hard Bayesian data fusion schemes developed in earlier work for collaborative human-robot sensing \cite{nisarTRO, lukeTRO, lukeRAS, Dani2015, doucette2012thesis, mehta2014}. This means that PSDA applies to non-Gaussian state distributions defined by mixture models, even when imperfect priors are available and computational approximations must be made for online fusion. Although PSDA requires additional semantic sensor characteristics like false positive rates, it is robust to modest mismatches between assumed and actual parameter values. These properties are extensively demonstrated in this work through simulations of a futuristic human-robot scientific planetary survey task. 

In the rest of this paper, Sec. \ref{sec: section_back} reviews semantic data fusion in the context of human-robot interaction and conventional DA.
Sec. \ref{sec: section_sda} formalizes the SDA problem statement and discusses potential solution approaches in relation to conventional DA methods. 
Sec. \ref{sec: section_psda} describes PSDA for single object state estimation, then generalizes PSDA to the multi-object case. As a special case, a variational method building on the approach of \cite{nisarTRO} is introduced for approximating a PSDA state posterior when the prior is a Gaussian mixture (GM) and the semantic observation likelihood is a softmax function. Sec. \ref{sec: section_sim} presents simulation results, and conclusions are presented in Sec. \ref{sec: section_conc}.


\section{Background} \label{sec: section_back}
In this section, semantic data and uncertainties for state estimation are defined. Focusing on soft semantic data that may contain incorrect or ambiguous references, relevant modeling and fusion techniques from human-robot/machine interaction and DA literature are reviewed. The shortcomings of these works motivate our formal SDA problem statement in the next section and the development of our PSDA solution. 
\subsection{Semantic Observations and Uncertainties} \label{sec:semobsuncs}
Semantic observations \textcolor{black}{are used here to refer to noisy categorical descriptions of abstract object/event properties or relations}. Such \textcolor{black}{observations} can for instance be provided by human collaborators via communication interfaces based on language or other interaction modalities (e.g. gestures), or by various perception algorithms such as object detectors\footnote{Although the motivation here is towards collaborative human-robot state estimation with soft (i.e. human-provided) data, the problem framing and methodology extends to situations involving only machine data sources.}. 
Throughout this work, the term `semantic \textcolor{black}{observations}' will restrictively refer to \textcolor{black}{semantic variables corresponding to} categorical descriptions of continuous states. 
\textcolor{black}{This usage is uncommon in robotics, but motivated by the usefulness of meaningfully grounded automated reasoning over continuous variables like spatial positions, velocities, masses, temperatures, terrain grades, etc. \cite{bishop2011, tse2018, Dani2015, doucette2012thesis, frost2010,zheng2021spatial, lukeRAS}}. 

{
Probabilistic models \textcolor{black}{for providing semantic observations} 
must capture the uncertain and imprecise nature of hybrid mappings from continuous states to discrete semantic labels. 
This can be done via classifier models \cite{nisar2018,tse2018}, neural networks \cite{Dani2015}, density function regression \cite{porta2006}, or grid-based representations \cite{frost2010, prendergast2018, zheng2021spatial}. For instance, such models can describe the conditional probability of observing a particular semantic preposition label of a target object's spatial location with respect to a reference object or frame, given a particular (hypothetical) continuous target state configuration (e.g. `the box is front of the robot, next to the wrench, between the street and building'). As discussed in \cite{lukeRAS, wyffels2015, koch}, they can also describe \textit{negative} semantic \textcolor{black}{observations}, 
which extract state information from an \textcolor{black}{anticipated} lack of 
positive observations (e.g. `nothing is inside the box', `landmark A not in view', `target is occluded').  
}


\begin{table}[]
\caption{\textcolor{black}{Common} semantic uncertainties in collaborative estimation \textcolor{black}{besides spatial uncertainty}}
\label{tab: classes_of_semantic_uncertainty}
\begin{tabular}{|l|l|llll}
\cline{1-2}
Type & Description &  &  &  &  \\ 

\cline{1-2} 1) Transmission clarity & 
\begin{tabular}[c]{@{}l@{}} If the phrase is provided via speech \\ in the presence of background noise or \\ a poor microphone, it may be \\ mistranslated by a speech processing \\ algorithm as `I 
see a good cow sight \\ specimen near the center of the creator'. \end{tabular} &  &  &  &  \\ 

\cline{1-2} \begin{tabular}[c]{@{}l@{}}2) Parsing and extraction of \\ relevant statement entities\end{tabular} & \begin{tabular}[c]{@{}l@{}} 
Whether the clause \\ `good 
calcite specimen' is understood as \\ a rock object, `center of  the crater' \\ is understood as a map feature, etc. \end{tabular} &  &  &  &  \\ 

\cline{1-2} 3) Lexical synonymy & \begin{tabular}[c]{@{}l@{}} 
Whether `near' is understood as being \\ similar to other terms, e.g. `close to', \\ `next to', `nearby', etc. \end{tabular} &  &  &  &  \\ 

\cline{1-2} \begin{tabular}[c]{@{}l@{}}4) Correctness and \\ target assignment\end{tabular} & \begin{tabular}[c]{@{}l@{}} 
Whether a calcite 
specimen \\ really was seen (or else another 
mineral \\ type rock was observed), and if so \\ whether it is of  `round' or `large' type \end{tabular} &  &  &  &  \\ \cline{1-2}
\end{tabular}
\vspace{-0.1in}
\end{table}

In practice, other types and sources of semantic uncertainty are also important when considering collaborative estimation. This can be illustrated by returning briefly to the simple example in Fig. \ref{fig:data_association_example}. For context, in this simplified planetary survey application (elaborated in Sec. \ref{sec: section_sim}), scientifically desirable rock specimens are either of type `round' (with few/no face edges) or `large' (of at least some diameter). Suppose the team searches for one of each type of specimen, so that there are two unknown but distinct location states of interest for a satisfactory `round' specimen and `large' specimen. Then, as described in Table \ref{tab: classes_of_semantic_uncertainty}, the uncertainty associated with the soft data observation `I see a good calcite specimen near the center of the crater' depends on several other factors (aside from the spatial uncertainty of `nearby the center of the crater').

For many applications, items (1)-(3) can be mitigated by template-based structured language interfaces and learning-based language models. For example, to address these concerns in spatial localization tasks, Sweet \cite{sweet2016, nick2016thesis} used free-form text interfaces to maximize transmission clarity, along with off the shelf language processing tools to generate suitable phrase parsings that mapped to manually-specified observation templates. Word sense-matching techniques \cite{mikolov2013} were also used to handle synonymous phrasings from free-form text, enabling selection of appropriate \textcolor{black}{spatial} semantic likelihood functions based on trained softmax models from a compact semantic descriptor dictionary. Sweet and Ahmed \cite{sweet2016, nisar2018} showed that such dictionaries can be learned with a minimal training data and reshaped to various contextual geometries via algebraic operations. Extensions are also possible for cases where reference frames (absolute, relative, intrinsic), environment maps, object groundings, and vocabularies are not necessarily known a priori \cite{lukeFUSION, howard2014, matuszek2013, zheng2021spatial}. 
Yet, these approaches are insufficient to address item (4), since they do not reason about the \textcolor{black}{consistency of 
spatial semantic information relative to other sensor data or state beliefs.} 

This work seeks to rigorously close this gap from a state estimation and data fusion perspective. We argue that items (1)-(3), along with uncertainties in reference frames, groundings and so on, can be logically set aside by \textit{conditioning} reasoning about correctness and target assignment on semantic interpretations available up to that point. That is, received semantic observations are assumed plausible, i.e. resolvable via pre-processing into well-formed \textit{candidate} hybrid likelihood functions that \textit{may} be ascribed to particular states. \footnote{This approach is also justifiable in view of the probabilistic semantic data fusion method proposed by Sweet \cite{sweet2016, nick2016thesis} - their approach marginalizes state estimates over distinct semantic observation interpretation hypotheses, where hypothesis probabilities can for instance be derived from output scores of template- and word sense-matchers.}   
\subsection{Assessing  Correctness and Target Assignment}
{Sophisticated probabilistic methods have been developed for robots to interpret commands provided by human supervisors and teammates, accounting for uncertainties in grounding and context 
inherent to natural language instructions \cite{tellex2011g3, matuszek2013, howard2014, duvallet2016}. 
For collaborative sensing, there are also many techniques for ingesting both structured and free-form human semantic observations within hybrid Bayesian data fusion and state estimation algorithms. 
} However, these approaches all assume that entities being referred to actually do exist in the environment, so humans are treated as oracles whose inputs definitely match objects of interest. Thus, while some techniques can use environment information and training data to resolve grounding/assignment ambiguities among objects known to exist, they cannot in general reason about possible false references to non-existent entities. 

Rosenthal, et al. \cite{rosenthal2011} considers a POMDP-based framework for human-assisted robotic planning in indoor navigation tasks that accounts for the accuracy of human sensors that provide discretized robotic location and guidance information over a known map. Wang, et al. \cite{wang2012} proposes a maximum likelihood framework to jointly perform state estimation and multiple human sensor parameter calibration for binary event data (learning of false alarm and missed detection rates). Ahmed et al. \cite{nisar2015sketch} and Muesing, et al. \cite{muesing2021} likewise consider hierarchical Bayesian inference and learning methods to address the same issues in sketch-based target localization and interactive dynamic target classification problems, respectively. While these works all attempt to rigorously account for the probability that human sensors may provide false soft data, they do not consider the problem of matching valid but ambiguous soft data to distinct states of interest. Furthermore, they all assume discrete state spaces and observation models, and thus do not accommodate the hybrid semantic measurement likelihoods considered here. Burks and Ahmed \cite{lukeFUSION, nisarTRO} attempted to address this gap heuristically using mixture models to account for assumed constant probabilities of erroneous and ambiguously matched semantic soft data. This provides a simple convenient way to `hedge against' these uncertainties, but requires careful parameter tuning of static parameters which cannot adapt to actual state beliefs. Available state information is thus blindly diluted, which can lead to overly conservative fusion results. 

Random finite set methods are also popular for semantic data fusion in human-machine interaction. Bishop and Ristic \cite{bishop2011} used random set theoretic likelihoods to account for multiple interpretations of semantic natural language sentences, owing to the ambiguity of spatial prepositions. 
Their method accounts for the possibility that a human sensor provides invalid observations, but does not use human sensor models or data derived from other sensors to perform dynamic {\it a posteriori} inference to update understanding about the correctness of semantic human data (as the approach developed here does). Also, their method does not disambiguate semantic data that might refer to one of many targets. 

\subsection{Data Association Methods} \label{sec: conventional_pda}
With these gaps in mind, we ideally seek techniques for dynamic {\it a posteriori} assessment of semantic observation correctness and target assignment.  
These properties correlate with solutions to the general problem of \textit{data association} (DA),  
which arises in state estimation whenver continuous hard sensor data is of uncertain origin or contains clutter/false signals. Such situations are common in traffic monitoring for aircraft and self-driving cars, where sensor signals are ambiguous as to which objects caused them and objects may be incorrectly tracked if data are incorrectly assigned to sources. 
For concise overviews of DA problems in robotics, see \cite{Neira2001, schulz2001}. 

\begin{figure}[t]
    \centering
    \includegraphics[width=5.25cm]{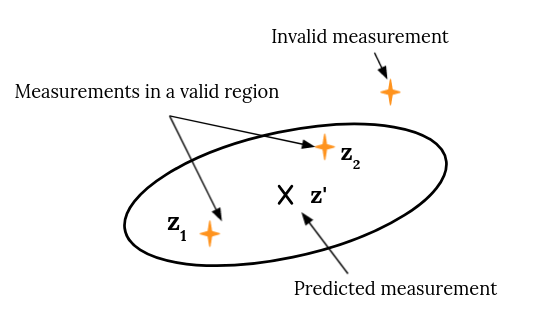}
    \caption{Illustration of conventional PDA for hard data fusion: the ellipse illustrates the validation gate; measurements outside the gate are ignored and association probabilities for each valid measurement ($z_1$ and $z_2$) are calculated for updating object state estimates.}
    \label{fig:conventional_pda} 
    \vspace{-0.2in}
\end{figure}
Many conventional DA methods have been developed, such as nearest neighbor \cite{kenari2014}, multi-hypothesis tracking (MHT) \cite{vo2015}, probabilistic data association (PDA) \cite{bar-shalom1975}, \cite{bar-shalom2009}, and particle filter-based  Monte Carlo data association \cite{sarkka2007}. These often require knowledge of the number of objects to be tracked. In recent years, techniques based on random finite set statistics like the probability hypothesis density (PHD) filter have also become popular for their ability to estimate the number of tracked objects and their states simultaneously, while accounting for clutter measurements (false positives) and semantic data \cite{mahler2007, dames2015}. This affords certain advantages over conventional DA, but introduces implementation issues, e.g. the need for ad hoc methods to extract object state covariances. As such, methods like PDA and MHT remain attractive for a known/bounded number of target objects, as assumed here. 
Yet, the assumptions underlying methods like nearest neighbor, MHT, and PDA prevent application to semantic data. 
For instance, to limit the data association hypothesis space to only feasible measurement-target pairs, conventional DA methods rely on Gaussian approximations to perform measurement gating on continuous sensor data, as shown in Fig. \ref{fig:conventional_pda}. Such gating has no obvious extension for discrete semantic data. More generally, since semantic data fusion requires operations with highly non-Gaussian state pdfs and hybrid likelihoods for discrete observations, typical DA algorithm computations cannot be directly applied. 
Thus, the DA problem and solutions must be suitably extended to semantic \textcolor{black}{observations}. 
%



\section{THE SDA PROBLEM} \label{sec: section_sda}
\begin{figure}[t]
    \centering
    \includegraphics[width=6.25cm]{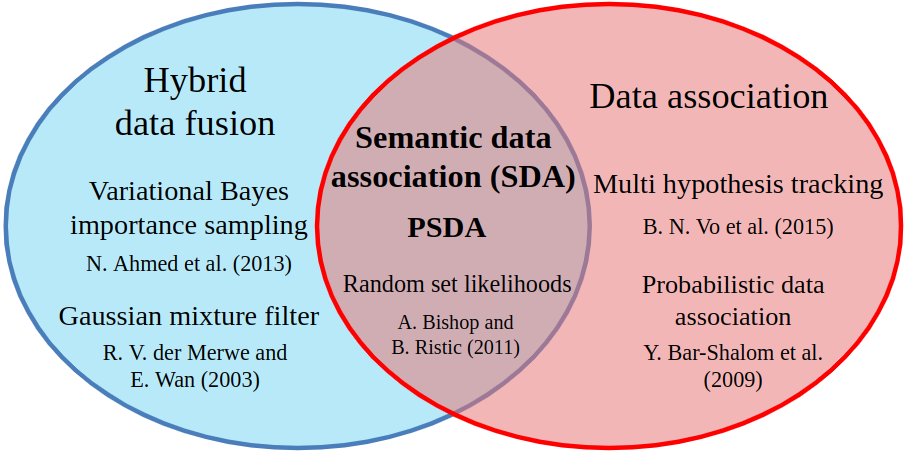}
    \caption{The semantic data association (SDA) problem resides at the intersection of data association and hybrid soft-hard fusion for semantic \textcolor{black}{observations}. The PSDA algorithm is one possible way to solve SDA probabilistically.}
    \label{fig: sda}
    \vspace{-0.2in}
\end{figure}

As illustrated in Fig. \ref{fig: sda}, this work focuses on the gap between DA and semantic data fusion, and labels the class of estimation problems that consider the correctness and target association of semantic \textcolor{black}{observations}, along with extraction of state information, as {\it semantic data association (SDA)}. There are few methods for SDA aside from \cite{bishop2011, mahler2007} which account for correctness, and none (to the best of our knowledge) that also address target association ambiguity. In this section, the SDA problem is first defined formally. Then possible solutions are considered based on theoretical similarities to existing DA algorithms. One such extension of PDA is then fully developed in the next section via the new {\it probabilistic semantic data association (PSDA)} algorithm. 
For simplicity, the problem is framed with respect to static states of interest for a single semantic human sensor and robot, but is easily generalized to dynamic states and multiple human sensors. 

\subsection{General Problem Statement} \label{sec: general_problem_statement}

\begin{figure}[t]
    \centering
    \includegraphics[width=7cm]{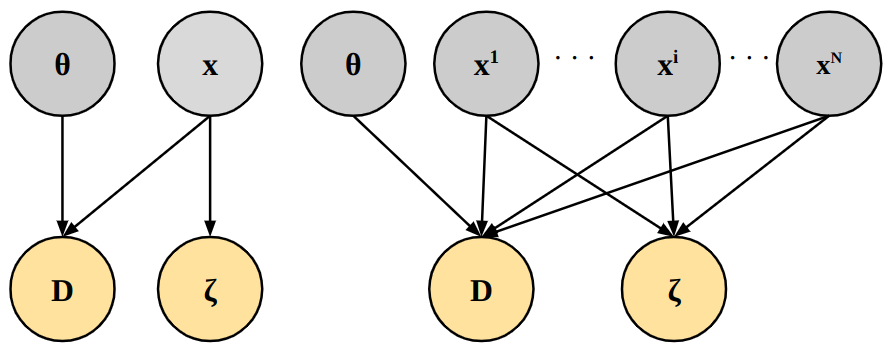}
    \caption{Graphical models for hybrid data fusion with ambiguous and potentially false semantic data for a single-(left) and multi-(right) state problem, at a single time step, with unobserved random variables shown in gray.}
    \label{fig:graphical_models}
    \vspace{-0.2in}
\end{figure}

Let $x_k$ be a particular state of interest 
at some time index $k \in \mathbb{Z}^{0+}$. Let $\zeta_k$ be the robot's sensor observation of the target at discrete time index $k$ and $D_k$ be a discrete variable representing a semantic observation of the target object given by a human collaborator at $k$. Following Sec. \ref{sec:semobsuncs}, let $D_k$ take label values indexed by $j \in \left\{1,\cdots, H \right\}$, where the labels are mutually exclusive and exhaustive for a dictionary describing a particular kind of semantic observation $D_k$ (i.e. $\sum_{j=1}^{H}p(D_k=j|x_k) = 1$). Also, let $\theta_k$ be a latent data association variable, which represents the index of an unknown \textit{correctness and target association hypothesis} (hereafter simply called an `association hypothesis') for semantic observation $D_k$. For example, if there is only a single object of interest with state $x_k$ (hereafter, the state of interest), then $\theta_k$ is either $0$ or $1$. If $\theta_k=0$, then $D_k$ is a false/incorrect description that describes no actual state (i.e. the human provided mistaken input), or equivalently, describes some other object state that has nothing to do with $x_k$ (some other object was mistaken by the human for the object of interest). If $\theta_k=1$, $D_k$ is a correct description of the object of interest and corresponds to $x_k$. More generally, when there are \textcolor{black}{$N$} distinct states of interest $x^i_k$ for $i=1,\cdots,\textcolor{black}{N}$, then $\theta_k$ can take any integer value from $0$ ($D_k$ does not correctly correspond to any of the \textcolor{black}{$N$} states) to \textcolor{black}{$N$} ($D_k$ correctly corresponds to $x^{\textcolor{black}{N}}_k$). Fig. \ref{fig:graphical_models} shows the corresponding graphical models, where 
distinct $D_k$ observations are assumed conditionally independent of each other given $x_k$ or all $x^i_k$ (this condition could be relaxed as in \cite{muesing2021} and \cite{nisar2015sketch}, but outside the scope of this work).

The goal of the SDA is to estimate the posterior distribution of the states of interest by fusing semantic data $D_k$ and robotic sensor data $\zeta_k$, while inferring the latent association variable $\theta_k$. From a general Bayesian estimation perspective, assuming some form of joint prior 
$p(x^1_k,\cdots,x^{\textcolor{black}{N}}_k, \theta_k)$ along with observation likelihood models for $D_k$ and $\zeta_k$, this requires finding the joint posterior 
{\allowdisplaybreaks
\abovedisplayskip = 2pt
\abovedisplayshortskip = 2pt
\belowdisplayskip = 0pt
\belowdisplayshortskip = 0pt
\begin{align}
&p(x^1_k,...,x^{\textcolor{black}{N}}_k, \theta_k | D_k, \zeta_k) \nonumber \\
&\propto p(D_k, \zeta_k | x^1_k,...,x^{\textcolor{black}{N}}_k, \theta_k) p(x^1_k,...,x^{\textcolor{black}{N}}_k, \theta_k). \label{eq:joint_SDA_hypstate_post}
\end{align}
}

\subsection{Possible Solution Approaches}
The type of algorithm needed to solve the SDA problem depends on how it is desired to handle $\theta_k$ in (\ref{eq:joint_SDA_hypstate_post}). For instance, a method which seeks a single \textcolor{black}{most probable} semantic association hypothesis $\theta_k$ by keeping semantic observations for a certain window of time and then comparing posterior distributions can be considered an extension of either nearest neighbor or MHT \cite{vo2015}. Alternatively, a method which seeks the marginal of (\ref{eq:joint_SDA_hypstate_post}) 
by probabilistically summing over all SDA hypotheses at each time, i.e.
{\allowdisplaybreaks
\abovedisplayskip = 2pt
\abovedisplayshortskip = 2pt
\belowdisplayskip = 0pt
\belowdisplayshortskip = 0pt
\begin{align}
&p(x^1_k,...,x^{\textcolor{black}{N}}_k| D_k, \zeta_k) = 
\sum_{\theta_k = 0}^{\textcolor{black}{N}}  p(x^1_k,...,x^{\textcolor{black}{N}}_k, \theta_k | D_k, \zeta_k) \nonumber \\
&= \sum_{\theta_k = 0}^{\textcolor{black}{N}}  p(x^1_k,...,x^{\textcolor{black}{N}}_k| \theta_k, D_k, \zeta_k) p(\theta_k| D_k, \zeta_k), \label{eq:decomppost}
\end{align}
}can be considered a theoretical extension of the PDA method \cite{bar-shalom1975, bar-shalom2009}. This 
 leads to the proposed PSDA algorithm. 

Note that according the multi-state graphical model in Fig. \ref{fig:graphical_models} (right), $\theta_k$ and all states $x^1,\cdots,x^{\textcolor{black}{N}}$ become conditionally dependent given $D_k$ and $\zeta_k$. Thus, estimation for $x^i$ alone must still account for uncertainties in all other states $i' \neq i$, which introduces subtle complexities to the problem. 

\section{THE PSDA ALGORITHM} \label{sec: section_psda}
To build up intuition, the PSDA algorithm is derived first from the single state perspective, which implies reasoning over the correctness of $D_k$ only. General multi-state PSDA is then presented, which reasons over both the correctness and assignment of $D_k$ to particular target states. Lastly, the issue of computing the (generally intractable) posteriors $p(x^1_k,...,x^{\textcolor{black}{N}}_k| \theta_k, D_k, \zeta_k)$ and $p(\theta_k|D_k,\zeta_k)$ in (\ref{eq:decomppost}) for hybrid semantic likelihoods $p(D_k|x_k)$ is considered, and an efficient approximation technique building off prior work is given for the special case of softmax likelihoods with GM priors. 

\subsection{Single state PSDA} \label{sec: pda_single_target} 
In the following, assume that the robot sensor data $\zeta_k$ is already fused into the state prior pdf before fusing $D_k$. For ease of notation, we also omit the time index $k$. 
Following the graphical model in Fig. \ref{fig:graphical_models} (left), let the joint prior $p(x,\theta) = p(x)p(\theta)$. 
The marginal posterior $p(x|D)$ is 
{\allowdisplaybreaks
\abovedisplayskip = 2pt
\abovedisplayshortskip = 2pt
\belowdisplayskip = 2pt
\belowdisplayshortskip = 0pt
\begin{align}
    p(x|D) &= \sum_{\theta} p(x, \theta|D) 
           = \sum_{\theta} p(x|D, \theta)p(\theta|D), \label{eq:second_expansion_single_target}
\end{align}
}
\noindent where the first term of 
is the posterior of a target object's state given $\theta$ and $D$, and the second term is the marginal posterior of $\theta$ given $D$. 
Applying Bayes' rule and marginal prior independence of $x$ and $\theta$, the posterior  $p(\theta|D)$ is 
{\allowdisplaybreaks
\abovedisplayskip = 2pt
\abovedisplayshortskip = 2pt
\belowdisplayskip = 2pt
\belowdisplayshortskip = 0pt
\begin{align}
    p(\theta|D) &= \frac{p(D|\theta)p(\theta)}{\sum_{\theta}p(D|\theta)p(\theta)} 
    = \frac{\int_x p(D, \theta, x)dx}{\sum_{\theta}\int_x p(D, \theta, x)dx} \nonumber \\
    &= \frac{p(\theta)\int_x p(D|\theta, x)p(x)dx}{\sum_{\theta}p(\theta)\int_x p(D|\theta, x)p(x)dx}. \label{eq:p_theta_given_D_single}
\end{align}
}
\noindent For $\theta=0$ and $\theta=1$, 
the single object PSDA posterior 
is 
{\allowdisplaybreaks
\abovedisplayskip = 2pt
\abovedisplayshortskip = 2pt
\belowdisplayskip = 2pt
\belowdisplayshortskip = 0pt
\begin{align}
    &p(x|D) = \nonumber \\
    & p(x) \frac{p(\theta=0)\int_xp(D|\theta=0, x)p(x)dx}{\sum_\theta p(\theta) \int_x p(D|\theta, x)p(x)dx} \nonumber \\
    + \ &\frac{p(D|x)p(x)}{\int_xp(D|x)p(x)dx} \frac{p(\theta=1)\int_xp(D|\theta=1, x)p(x)dx}{\sum_\theta p(\theta) \int_x p(D|\theta, x)p(x)dx} \label{eq:posterior_binary_association},
\end{align}
}
\noindent where $p(D|\theta=0, x)$ is independent of $x$, so that 
{\allowdisplaybreaks
\abovedisplayskip = 2pt
\abovedisplayshortskip = 2pt
\belowdisplayskip = 2pt
\belowdisplayshortskip = 0pt
\begin{align}
    &p(D|\theta=0, x) = \frac{p(D, \theta=0, x)}{p(\theta=0, x)} \nonumber \\ 
    &= \frac{p(x|D, \theta=0)p(D, \theta=0)}{p(\theta=0, x)} \nonumber 
    = \frac{p(x)p(D, \theta=0)}{p(x)p(\theta=0)} \nonumber \\
    &= \frac{p(D, \theta=0)}{\sum_D p(D, \theta=0)}.
\end{align}
}
Here we assume that the value of $p(D, \theta=0)$ is uniform {regardless of which discrete integer value $D$ takes,} so that $p(D|\theta=0, x) = 1/H$, where $H$ is the number of possible labels in the semantic dictionary for $D$'s observation type. 
From (\ref{eq:posterior_binary_association}), it can therefore be said that the {single object PSDA} posterior $p(x|D)$ is a mixture of two pdfs: the prior $p(x)$ (i.e. the posterior under $\theta=0$) and the conditional posterior, 
{\allowdisplaybreaks
\abovedisplayskip = 2pt
\abovedisplayshortskip = 2pt
\belowdisplayskip = 2pt
\belowdisplayshortskip = 0pt
\begin{align}
p(x|D,\theta=1) &= \frac{p(D|x,\theta=1)p(x)}{\int_x p(D|x,\theta=1)p(x)dx} \nonumber \\
&= \frac{p(D|x)p(x)}{\int_xp(D|x)p(x)dx}, \label{eq:condpost_single}
\end{align}
}
which are each weighted by association probabilities $p(\theta|D)$. 

This reasoning over the posterior for $\theta$ is the same basis for the conventional PDA algorithm. 
However, as detailed later in this section, we do not need to assume Gaussian prior pdfs or measurement likelihoods as in conventional PDA, and therefore can handle non-Gaussian priors/posteriors and multi-modal semantic likelihoods.  
Also unlike conventional PDA, gating is usually of little concern, as the volume of semantic data to be fused at a single time step is typically much lower for soft data fusion than for conventional sensors like lidar or radar. 
More general state-dependent ``semantic clutter'' models could also be used in place of the uniform distribution for $p(D,\theta=0,x)$, as sometimes done in conventional PDA. 
These similarities and differences 
 are elaborated further in the next subsection, for the general multi-object case. 


\subsection{Multi-state PSDA}
In this case, the latent data association variable $\theta_k$ can take one of $\textcolor{black}{N} + 1$ integer values where $\textcolor{black}{N}$ is the total number of possible 
target objects.  
The right graphical model in Fig. \ref{fig:graphical_models} represents this situation for a single time step. 
In the following, as with a single target object case, the time index $k$ is omitted, and denote $D = j$ as $D_j$ and $\theta = i$ as $\theta_i$ for $i \in \{0, \cdots, \textcolor{black}{N}\}$ for brevity, where $0$ denotes `false data'. Again, let robot data $\zeta$ already be fused into the prior. 

Let  $X = [x^1, \cdots, x^{\textcolor{black}{N}}]^T$ represent the combined state vector with prior $p(X) = p(x^1,\cdots,x^{\textcolor{black}{N}}) = \prod_{i=1}^{\textcolor{black}{N}}p(x^i)$. Then the joint posterior $p(X|D_j)$ is 
{\allowdisplaybreaks
\abovedisplayskip = 2pt
\abovedisplayshortskip = 2pt
\belowdisplayskip = 2pt
\belowdisplayshortskip = 0pt
\begin{align}
    p(X|D_j) &= \sum_{i=0}^{\textcolor{black}{N}}p(X, \theta_i|D_j) 
    = \sum_{i=0}^{\textcolor{black}{N}}p(X|D_j, \theta_i)p(\theta_i|D_j) \nonumber \\
    &= \sum_{i=0}^{\textcolor{black}{N}}p(X|D_j, \theta_i)\frac{p(D_j, \theta_i)}{\sum_s p(D_j, \theta_s)}. \label{eq:joint_D_theta_multi}
\end{align}
}
Following the detailed derivations provided in Appendix \ref{appx: derivation}, it can be shown that eq. (\ref{eq:joint_D_theta_multi}) reduces to
{\allowdisplaybreaks
\abovedisplayskip = 2pt
\abovedisplayshortskip = 2pt
\belowdisplayskip = 2pt
\belowdisplayshortskip = 0pt
\begin{align} 
    &p(X|D_j) =  
     p(X) \frac{\frac{1}{H}p(\theta_0)}{\frac{1}{H}p(\theta_0) + \sum_{s=1}^{\textcolor{black}{N}} p(D_j, \theta_s)} \\
    & + \sum_{i=1}^{\textcolor{black}{N}} \Big\{ p(x^i|D_j, \theta_i) \prod_{l \neq i} p(x^l) \Big\} \frac{p(D_j, \theta_i)}{\frac{1}{H} p(\theta_0) + \sum_{s=1}^{\textcolor{black}{N}} p(D_j, \theta_s)},
    \label{eq: joint_posterior_general}
\end{align}
}
\noindent where $p(\theta_0)$ {denotes the false detection} rate of the semantic sensor providing $D$. {For collaborative human-robot sensing, $p(\theta_0)$ will vary from individual to individual. As in the single target case, it can be seen that the multi-state PSDA posterior is a mixture model, though the number and complexity of the non-`false data' association hypothesis terms have increased.} 

The term $p(D_j, \theta_i)$ can be expanded as $\int_{x^i} p(D_j|\theta_i, x^i) p(x^i) p(\theta_i)dx^i$, so that the mixture weights {$\gamma_{PSDA}$} (i.e. association probabilities) become 
{\allowdisplaybreaks
\abovedisplayskip = 2pt
\abovedisplayshortskip = 2pt
\belowdisplayskip = 2pt
\belowdisplayshortskip = 0pt
\begin{align}
\gamma_{PSDA, i} = 
\begin{cases}
\frac{\frac{1}{H}p(\theta_0)}{\frac{1}{H}p(\theta_0) + \sum_{s=1}^{\textcolor{black}{N}} p(D_j, \theta_s)} & (i=0)  \\
\frac{p(D_j, \theta_i)}{\frac{1}{H} p(\theta_0) + \sum_{s=1}^{\textcolor{black}{N}} p(D_j, \theta_s)} & (i \neq 0) 
\end{cases}
\label{eq: gamma_0i}
\end{align}
}

\noindent Finally, each 
target \textcolor{black}{PSDA} posterior $p(x^i|D_j)$ is derived by marginalizing $p(X|D_j)$ with respect to all but index $i$, 
{\allowdisplaybreaks
\abovedisplayskip = 2pt
\abovedisplayshortskip = 2pt
\belowdisplayskip = 2pt
\belowdisplayshortskip = 2pt
\begin{align} 
    &p(x^i|D_j) = \int_{x^{l\neq i}}p(X|D_j)dx^{l\neq i} \nonumber \\
    &= {\gamma_{PSDA, i}} \cdot  p(x^{i,*}) + \Big\{ \sum_{l=0, l \neq i} {\gamma_{PSDA, l}} \cdot p(x^i) \Big\}, \label{eq:final_psda_posterior}
\end{align}
}
where $p(x^{i,*})$ is the updated posterior of $x^i$ using $D_j$.

\subsubsection{Comparison of conventional PDA and PSDA}
Here, we further discuss the connection between conventional PDA and PSDA algorithms. The details of the derivation of the conventional PDA algorithm are given in \cite{bar-shalom2009}. PDA is based on Kalman filter (KF) {\cite{kalman1960}} when the state transition and measurement models of the target object are linear, or linearized filters such as the extended Kalman filter (EKF){\cite{julier1997}} when they are nonlinear, so that the association probability ${\gamma_{PDA, i}}$ for each of $m$ valid continuous measurements $z_i$, (such as shown in Fig. \ref{fig:conventional_pda}) becomes
{\allowdisplaybreaks
\abovedisplayskip = 2pt
\abovedisplayshortskip = 2pt
\belowdisplayskip = 2pt
\belowdisplayshortskip = 2pt
\begin{align}
\gamma_{PDA, i} =
\begin{cases}
\frac{1-p_d p_g}{1- p_d p_g + \sum_{j=1}^{m} \mathcal{L}_j} & (i=0), \\
\frac{\mathcal{L}_i}{1- p_d p_g + \sum_{j=1}^{m} \mathcal{L}_j} & (i = 1, \cdots, m),
\end{cases}
\label{eq: conv_data_association_prob}
\end{align}
}


\noindent where $p_d$ is the target detection probability, $p_g$ is a gate probability (i.e. probability that a validation region contains a true measurement if detected, cf. Fig. \ref{fig:conventional_pda}), $m$ is the total number of valid measurements and $\mathcal{L}_i$ is the ratio of the continuous measurement $z_i$ originating from the target rather than from clutter \cite{bar-shalom2009}. 
The values of $\gamma_{PSDA}$ in (\ref{eq: gamma_0i}) are the counterparts of  (\ref{eq: conv_data_association_prob}). The posterior of a tracked object state given continuous measurement $z_i$ in PDA is 
{\allowdisplaybreaks
\abovedisplayskip = 2pt
\abovedisplayshortskip = 2pt
\belowdisplayskip = 2pt
\belowdisplayshortskip = 0pt
\begin{align} \label{eq: conv_pda_posterior}
    p(x|z) = {\gamma_{PDA, 0}} \cdot p(x) + \sum_{i=1}^{m} {\gamma_{PDA, i}} \cdot p(x|z_i, \theta_i),
\end{align}
}
\noindent where $p(x)$ is a prior pdf of the target object. By comparing (\ref{eq: conv_pda_posterior}) with the PSDA posteriors in single-/multi- object cases, our expressions assume that the detection rate  $p_d$ is $1$ (i.e. no missed detection). The reason why (\ref{eq: conv_pda_posterior}) is a single posterior, while (\ref{eq: joint_posterior_general}) is a joint posterior, is that, in the case of discrete semantic observations, there is uncertainty about `which object the semantic observation was generated from' instead of `which continuous measurement is associated with an object', as with conventional PDA. Thus, in a more general light, PSDA apparently resembles the conventional joint PDA (JPDA) algorithm, which handles multiple targets in clutter and must also account for joint target state dependencies induced by DA uncertainty. Yet, PSDA is in fact more analogous to the so-called mixture PDA (MXPDA) approach \cite{salmond1990}, which represents the posterior of tracked targets via GMs.

\subsubsection{Greedy PSDA}
In the proposed PSDA method, after calculating all {$\gamma_{PSDA}$} terms, the pdfs corresponding to these values are combined to generate a new posterior. Since this new posterior will generally be used to form the prior for the next fusion instance, it is therefore generally desirable to manage the complexity of the mixture posterior in (\ref{eq:final_psda_posterior}) by compressing or truncating terms whenever possible to trade accuracy for computational efficiency and speed. One way to further simplify the results of the data association process and reduce computational cost of \textit{future} fusion calculations is by using only the association hypothesis pdf with the maximum association probability {as the new posterior.} 
For example, when ${\gamma_{PSDA}} = [{\gamma_{PSDA, 0}} = 0.1, {\gamma_{PSDA, 1}} = 0.5,{\gamma_{PSDA, 2}} = 0.4]^T$, only the pdf related to {$\gamma_{PSDA, 1}$} is set as a new posterior. We call this ``winner take all'' association method \textit{greedy-PSDA}, which can also be seen as a semantic data analog of the nearest-neighbor algorithm for continuous measurement DA. The estimation performance of greedy-PSDA is compared with that of the full PSDA algorithm in the next section. Note that for a single fusion update on the same prior and observation, the computational cost of greedy-PSDA is the same as full PSDA, since greedy-PSDA must still compute the same association weights in (\ref{eq: gamma_0i}). That is, the cost savings for greedy-PSDA are realized \textit{following} a fusion update, rather than during. Mixture compression \cite{salmond1990,runnalls2007} could also be used to the same end, but incurs higher immediate overhead.

\subsubsection{Special case: mixture priors and softmax likelihoods} \label{sec: special_case}
The hypothesis-conditional pdfs and weights for the mixtures in (\ref{eq:posterior_binary_association})  and (\ref{eq:final_psda_posterior}) are analytically intractable except for special choices of the hybrid semantic observation likelihood models discussed in Sec. \ref{sec:semobsuncs} (e.g. if a mixture of unnormalized Gaussians is used, then only certain types of prior pdfs for $x$ or $X$ will admit closed-form integrals). 
Therefore, calculations for PSDA will generally require approximations for these terms. 
Building off prior work in \cite{nisarTRO}, we show that computationally efficient approximations are readily available in the special case of softmax likelihoods and Gaussian mixture (GM) state prior pdfs. As discussed in \cite{nisar2018,lukeTRO, lukeRAS}, softmax models have a number of advantages over other semantic data likelihood models like \textit{unnormalized} GM functions 
(particularly with respect to scalability and learnability). Furthermore, softmax models support computations with GM pdfs, which approximate a wide variety of non-Gaussian priors and mesh well with PSDA's mixture-based computations. 

For PSDA, softmax models for the observation likelihood of $D_k=j$ with respect to a single state $x$ under the hypothesis $\theta=1$ are of the form
{\allowdisplaybreaks
\abovedisplayskip = 2pt
\abovedisplayshortskip = 2pt
\belowdisplayskip = 2pt
\belowdisplayshortskip = 0pt
\begin{align} \label{eq:softmax}
    p(D_k = j|x, \theta=1) = \frac{e^{w_j^T x_k + b_j}}{\sum_{h=1}^H e^{w_h^T x_k + b_h}},
\end{align}
}
\noindent where $w_j, w_h \in \mathbb{R}^n$ and $b_j, b_h \in \mathbb{R}^1$ are vector weight and scalar bias parameters, respectively. Each $\{w_j, b_j\}$ parameter pair determines the shape of $p(D_k=j|x_k, \theta=1)$ as a function of $x$ relative to all $H$ class categories. These parameters can be learned from calibration data and can also embed geometric invariance constraints, which are especially useful for spatial semantics \cite{nisar2018}. 
%
{As mentioned above, the association hypothesis conditional posterior $p(x|D,\theta=1)$ and conditional normalizing constant $p(D|\theta=1)=\int_x p(D|\theta=1, x)p(x)dx$ 
} cannot be computed analytically with this choice of likelihood. Although it is possible to approximate these expressions with a particle filter \cite{arulumpalam2002}, it is not necessarily robust for online implementation since 
a very large number of samples may be needed when $x$ is high-dimensional or if surprising semantic measurements are received which have low marginal probability $p(D|\theta=1)$. {Therefore, variational Bayes importance sampling (VBIS) method developed in \cite{nisarTRO} is used to approximate the \textcolor{black}{(marginalized)} PSDA posterior\textcolor{black}{, i.e.  (\ref{eq:posterior_binary_association}) for a single state or (\ref{eq:final_psda_posterior}) for multiple states}. This uses the fact that log-sum-exp function can be upper bounded \cite{bouchard2007},  
{\allowdisplaybreaks
\abovedisplayskip = 2pt
\abovedisplayshortskip = 2pt
\belowdisplayskip = 2pt
\belowdisplayshortskip = 0pt
\begin{align} 
    \log \Big(\sum_{h=1}^H e^{y_h} \Big) &\leq \alpha + \sum_{h=1}^H \frac{y_h - \alpha - \xi_h}{2} \nonumber \\ &+\lambda(\xi_h)\Big[(y_h - \alpha)^2 - \xi_h^2 \Big] + \log(1+e^{\xi_h}),
\end{align}}
\noindent where $\lambda(\xi_h) = \frac{1}{2\xi_h} \Big[\frac{1}{1+e^{-\xi_h}} - \frac{1}{2} \Big]$ and $y_h = w_h^T x + b_h$. Here, $\alpha$ and $\xi_h$ are free variational parameters, and the tightest upper bound of the log-sum-exp is obtained by optimizing these. Replacing the denominator of (\ref{eq:softmax}) with this bound, the softmax likelihood is approximated by a variational Gaussian $G(D, x)$. Thus, when the state prior distribution $p(x)$ is set as a Gaussian, the posterior can also be approximated as another Gaussian.} 
Although the mean of the hypothesis-conditional variational Bayes state posterior is often close enough to the true mean of $p(x|D,\theta=1)$, the approximated normalization constant {$\hat{C}_{VB} = \int_x p(x)G(D, x)dx$} is smaller than true 
constant $C = \int_x p(x)p(D|x,\theta=1)dx$ via the upper bound, 
{\allowdisplaybreaks
\abovedisplayskip = 2pt
\abovedisplayshortskip = 2pt
\belowdisplayskip = 2pt
\belowdisplayshortskip = 0pt
\begin{align*}
    C = \int_{x} p(x)p(D|x,\theta=1) dx \geq \int_x p(x) G(D, x) dx = \hat{C}_{VB}.
\end{align*}
}
This makes the approximated conditional posterior state covariance estimated by VB smaller than the true conditional posterior covariance, resulting in overconfident estimates. 

To address this, {Ahmed, et al. \cite{nisarTRO}} use importance sampling (IS) to improve the variational posterior approximation. IS can approximate the expected value of an arbitrary function $f(x)$ over a posterior pdf that is difficult to sample directly, by utilizing a Gaussian importance distribution $q(x)$ that is roughly similar in shape to $p(x|D,\theta=1)$. 
In particular, the IS approximation is constructed as 
{\allowdisplaybreaks
\abovedisplayskip = 2pt
\abovedisplayshortskip = 2pt
\belowdisplayskip = 2pt
\belowdisplayshortskip = 0pt
\begin{align*}
    \langle f(x) \rangle \approx \sum_{s=1}^{N_s} \omega_s f(x_s), \ \omega_s \propto \frac{p(x_s) p(D|x_s, \theta=1)}{q(x_s)},
\end{align*}
}
\noindent where $N_s$ is the number of samples, $\omega_s$ is the importance weight for sample $s$, 
and $q(x)$ is defined as 
{\allowdisplaybreaks
\abovedisplayskip = 2pt
\abovedisplayshortskip = 2pt
\belowdisplayskip = 2pt
\belowdisplayshortskip = 0pt
\begin{align} \label{eq: proposal_distribution}
    q(x) = {\cal N}(\mu_{VB}, \Sigma_{prior}),
\end{align}
}
where $\Sigma_{prior}$ is the covariance matrix of $p(x)$. The desired estimated mean and covariance for a Gaussian approximation of $p(x|D,\theta=1)$ are then $\hat{\mu}_{VBIS} = \langle x \rangle$ and $\hat{\Sigma}_{VBIS} = \langle (x-\mu)(x-\mu)^T\rangle$. {Moreover, since $\hat{C}_{VBIS} = \frac{\omega_s}{N_s}$, the required association hypothesis probability is readily obtained as a natural byproduct of approximating the conditional posterior $p(x|D,\theta=1)$ with VBIS.} 

In practical scenarios {that rely on recursive Bayesian updates}, the assumption that a prior $p(x)$ is Gaussian is usually violated, since {semantic data updates typically lead to multi-modal or heavily skewed state pdfs}. In such cases, Gaussian mixture (GM) pdfs provide useful and expressive prior models
{\allowdisplaybreaks
\abovedisplayskip = 2pt
\abovedisplayshortskip = 2pt
\belowdisplayskip = 2pt
\belowdisplayshortskip = 0pt
\begin{align} \label{eq: mixture_prior}
    p(x) = \sum_{u=1}^{M} w_u \cdot p(x|u),
    \vspace{-0.1in}
\end{align}
}
\noindent where $w_u \in \mathbb{R}^{0+}$ is a component weight and satisfies $\sum_{u=1}^M w_u = 1$. The VBIS procedure for a single Gaussian prior and softmax likelihood introduced above can be extended as described in Algorithm \ref{alg: vbis_gm} to a GM prior and multimodal sofmax likelihood to compute a GM posterior (see \cite{nisarTRO} for full details). {For the general case of multi-state PSDA, when the prior for state $x^i$ is an $M_i-$term GM, $p(x^i|D_j, \theta_i) \prod_{l \neq i} p(x^l)$ in (\ref{eq: joint_posterior_general}) can be further transformed as} 

\begin{algorithm}[t] 
\caption{VBIS for GMs} \label{alg: vbis_gm}
{\footnotesize
\begin{algorithmic}[1] 
\renewcommand{\algorithmicrequire}{\textbf{Input:}}
\renewcommand{\algorithmicensure}{\textbf{Output:}}
\REQUIRE {\it M}-component GM prior (\ref{eq: mixture_prior}), a semantic observation $D$, number of samples for importance sampling $N_s$  
\ENSURE posterior GM
\FOR {each mixture}
\STATE obtain initial estimates $\hat{\mu}_{VB}$ and $\hat{\Sigma}_{VB}$ via local VB(EM) update \cite{nisarTRO} 
\STATE draw $N_s$ samples from a proposal distribution (\ref{eq: proposal_distribution})
\STATE calculate an importance weight $\omega_s$ for each sample 
\STATE calculate normalizing constant $\hat{C}_{VBIS}$ as $\sum \omega_s / N_s$  
\STATE renormalize weights $\omega_s$
\STATE recalculate mean 
$\hat{\mu}_{VBIS}$ and covariance $\hat{\Sigma}_{VBIS}$
\ENDFOR
\end{algorithmic} 
}
\end{algorithm} 

{\allowdisplaybreaks
\abovedisplayskip = 2pt
\abovedisplayshortskip = 2pt
\belowdisplayskip = 2pt
\belowdisplayshortskip = 0pt
\begin{align} 
    &p(x^i|D_j, \theta_i) \prod_{l \neq i} p(x^l) 
    = \Big\{ \sum_{u^i=1}^{M_i}p(x^i, u^i|D_j, \theta_i) \Big\}\prod_{l \neq i} p(x^l) \nonumber \\
    &= \Big\{ \sum_{u^i=1}^{M_i}p(x^i|D_j, u^i)p(u^i|D_j) \Big\}\prod_{l \neq i} p(x^l), \label{eq:joint_states_posterior}
\end{align}
where $u^i$ is a latent variable to index the posterior GM mixand for state $x^i$, and 
{\allowdisplaybreaks
\abovedisplayskip = 2pt
\abovedisplayshortskip = 2pt
\belowdisplayskip = 2pt
\belowdisplayshortskip = 0pt
\begin{align} \label{eq: each_posterior_x_multi}
    p(x^i|D_j, u^i) = \frac{p(D_j|x^i)p(x^i|u^i)}{\int_{x^i}p(D_j|x^i)p(x^i|u^i)dx^i},
\end{align}
}
{\allowdisplaybreaks
\abovedisplayskip = 2pt
\abovedisplayshortskip = 2pt
\belowdisplayskip = 2pt
\belowdisplayshortskip = 0pt
\begin{align} \label{eq: each_posterior_u_multi}
    p(u^i|D_j) = \frac{p(u^i)\int_{x^i}p(D_j|x^i)p(x^i|u^i)dx^i}{\sum_{u^i=1}^{M_i}p(u^i)\int_{x^i}p(D_j|x^i)p(x^i|u^i)dx^i}.
\end{align}
}

\noindent In a similar manner, $p(D_j, \theta_i)$ in (\ref{eq: joint_posterior_general}) becomes 
{\allowdisplaybreaks
\abovedisplayskip = 2pt
\abovedisplayshortskip = 2pt
\belowdisplayskip = 2pt
\belowdisplayshortskip = 0pt
\begin{align} 
    p(D_j, \theta_i) &= \int_{x^i} p(D_j, \theta_i, x^i) dx^i \nonumber \\ 
    &= \int_{x^i} \sum_{u^i=1}^{M_i} p(D_j, \theta_i, x^i, u^i) dx^i \nonumber \\
    &= \int_{x^i}\sum_{u^i=1}^{M_i}p(D_j|x^i)p(x^i|u^i)p(u^i)dx^i. \label{eq:d_j_theta_i_multi}
\end{align}
\noindent 
By substituting (\ref{eq:d_j_theta_i_multi}) into (\ref{eq: gamma_0i}), association probabilities {$\gamma_{PSDA, 0}$} and {$\gamma_{PSDA, i}, \ i \in \{1, \cdots, \textcolor{black}{N} \}$} become 
{\allowdisplaybreaks
\abovedisplayskip = 2pt
\abovedisplayshortskip = 2pt
\belowdisplayskip = 2pt
\belowdisplayshortskip = 0pt
\begin{align} 
    &{\gamma_{PSDA, 0}} = \\ &\frac{\frac{1}{H}p(\theta_0)}{\frac{1}{H}p(\theta_0) + \sum_{s=1}^{\textcolor{black}{N}}\sum_{u^s=1}^{M_s} p(u^s)\int_{x^s}p(D_j|x^s)p(x^s|u^s)dx^s}, \label{eq: gamma_0_gm}
\end{align}
}
{\allowdisplaybreaks
\abovedisplayskip = 2pt
\abovedisplayshortskip = 2pt
\belowdisplayskip = 2pt
\belowdisplayshortskip = 0pt
\begin{align} 
    &{\gamma_{PSDA, i}} = \nonumber \\
    &\frac{\sum_{u^i=1}^{M_i} p(u^i)\int_{x^i}p(D_j|x^i)p(x^i|u^i)dx^i}{\frac{1}{H}p(\theta_0) + \sum_{s=1}^{\textcolor{black}{N}}\sum_{u^s=1}^{M_s} p(u^s)\int_{x^s}p(D_j|x^s)p(x^s|u^s)dx^s}. \label{eq: gamma_i_gm}
\end{align}
}
Therefore, to compute (\ref{eq: joint_posterior_general}), the equations (\ref{eq: each_posterior_x_multi}), (\ref{eq: each_posterior_u_multi}), (\ref{eq: gamma_0_gm}), and (\ref{eq: gamma_i_gm}) must be computed, which may seem complicated at first glance. However, for softmax likelihoods $p(D_j|x^i,\theta_i)$, the normalization constants needed to derive (\ref{eq: each_posterior_u_multi}), (\ref{eq: gamma_0_gm}), and (\ref{eq: gamma_i_gm}) are naturally obtained via the VBIS approximation to (\ref{eq: each_posterior_x_multi}). Thus, the added computational cost is negligible, as illustrated in Algorithm \ref{alg: gamma_gm}. In the simulations of Sec. \ref{sec: section_sim}, this VBIS method is applied when ``surprising'' positive measurements are obtained - that is, in cases where a conditional state posterior is expected to be very different from its corresponding state prior. 
When the shapes of a prior and a posterior are similar (i.e. $p(D|\theta_i)$ is not too small), likelihood weighted importance sampling (LWIS) \cite{bishop2006} is instead used with sampling weight 
{\allowdisplaybreaks
\abovedisplayskip = 2pt
\abovedisplayshortskip = 2pt
\belowdisplayskip = 2pt
\belowdisplayshortskip = 0pt
\begin{align}
    \omega_s \propto p(D|x_s), \ \  \forall s=1,\cdots, N_s
\end{align}
\noindent and $\mu_{LWIS}$ and $\Sigma_{LWIS}$ are computed by taking weighted sample averages. This is often the case, for instance, when persistent negative semantic information is received in search problems where the target is expected \textit{not} to be visible to sensors most of the time (e.g. $D = \mbox{`No  calcite  is near the center of the crater.'}$). 

\begin{algorithm}[t]
\caption{Calculation for the values of $\gamma_{PSDA}$} \label{alg: gamma_gm}
{\footnotesize
\begin{algorithmic}[1]
\renewcommand{\algorithmicrequire}{\textbf{Input:}}
\renewcommand{\algorithmicensure}{\textbf{Output:}}
\REQUIRE GM state priors, total number of possible observations $H$, false positive rate $FP (=p(\theta_0))$, semantic observation $D$
\ENSURE association hypothesis probabilities {$\gamma_{PSDA}$} 
\STATE initialize $GM\_\hat{C}\_arr$
\FOR {$s = 1: \textcolor{black}{N}$}
\FOR {$u = 1: M_s$}
\STATE $\mu_{VBIS}^s[u]$, $\Sigma_{VBIS}^s[u]$, $\hat{C}_{VBIS}^s[u]$ via Algorithm $1$
\ENDFOR
\STATE $GM\_\hat{C}\_arr[s] = \sum_{u=1}^{M_s} ( w_{prior}^s[u] \times \hat{C}_{VBIS}^s[u]$)
\ENDFOR
\STATE $\gamma_{den} = \frac{1}{H} FP + \sum_{s=1}^{\textcolor{black}{N}} GM\_\hat{C}\_arr[s]$ 
\STATE {$\gamma_{PSDA, 0} = (\frac{1}{H}FP) / \gamma_{den}$}
\FOR {$i = 1:\textcolor{black}{N}$} 
\STATE {$\gamma_{PSDA, i} = GM\_\hat{C}\_arr[i] / \gamma_{den}$}
\ENDFOR 
\end{algorithmic}
}
\end{algorithm}

\begin{figure*}[htbp]
    \centering
    \includegraphics[width=13.5cm]{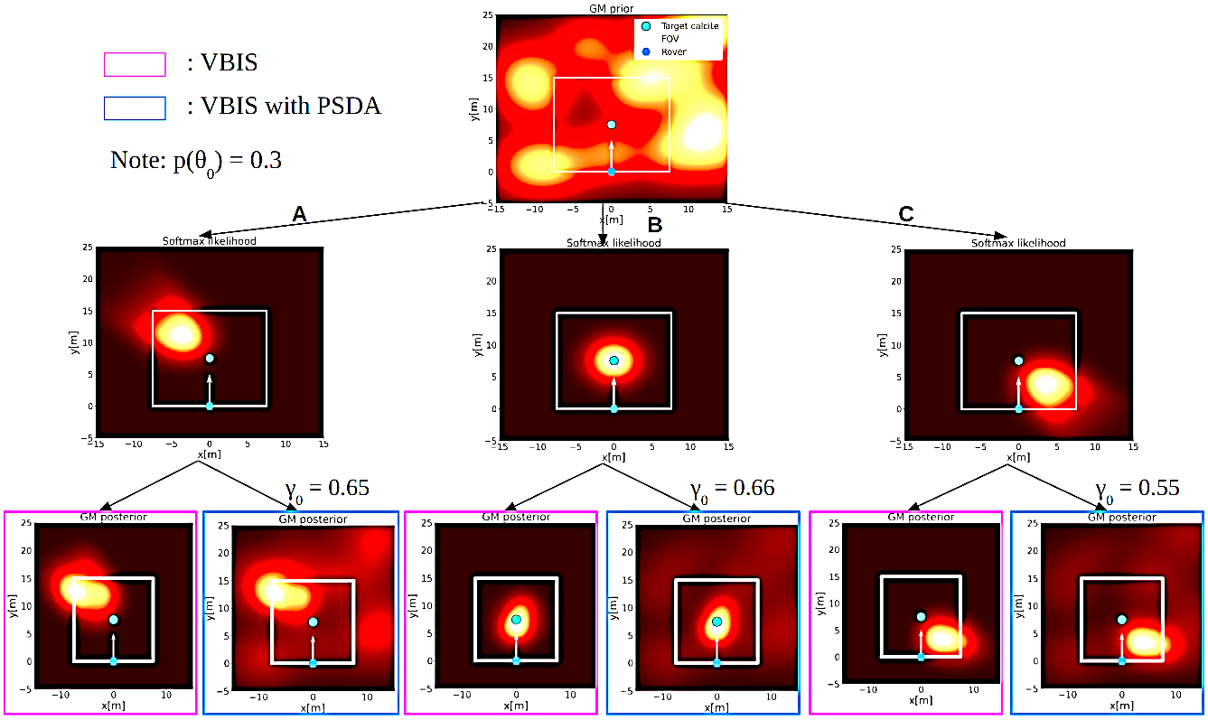}
    \caption{ (best viewed in color) GM pdfs (bottom row) for VBIS updates with PSDA (blue) and without (magenta) when different semantic observations (sofmax likelihoods in middle row) are provided for same GM prior (top), A: `Some good calcite is far left of you', B: 'Some good calcite is near ahead of you', and C: `Some good calcite is next to your right'; brighter values denote higher probability (blue dot is ground truth, white is rover FOV).}
    \label{fig:different_softmax_GMposterior} 
    \vspace{-0.2in}
\end{figure*}


\subsubsection*{Illustrative 2D Example} 
To demonstrate VBIS and PSDA, suppose the scientist in Fig.\ref{fig:data_association_example} 
provides different soft semantic data to the rover in a 2D search space while searching for a rounded calcite specimen (single state). The rover starts with a GM prior for the location of a desirable rounded calcite specimen (Fig. \ref{fig:different_softmax_GMposterior}, top), and uses VBIS with PSDA (bottom blue boxes) and VBIS without PSDA (bottom magenta boxes) to update the prior based on different received data, with $\gamma_0$ `false data' hypothesis posterior probability values shown for each PSDA update. As seen in Fig. \ref{fig:different_softmax_GMposterior}, in case B where the scientist provides correct data, 
the posterior pdf concentrates around the true calcite location both with and without PSDA.
In cases A and C where the scientist gives erroneous data, 
there is almost no probability around the target calcite when VBIS is used without PSDA. The pdf recovers around the true location via PSDA in these two cases, since the prior contributes to the posterior with probability $\gamma_0$. 
Whereas the most likely (MAP estimate) calcite location would be accurate in case B, they would be biased in cases A and C, resulting in wasted time and energy for rover path planning. However, it is possible to quickly recover better estimates in cases A and C with PSDA using more data, whereas this becomes much more difficult without PSDA. 
Extensive simulations presented next show that PSDA 
provides this robustness even when approximations like VBIS and GM compression are used repeatedly. This example also shows how $\gamma_{PSDA}$ depends on the agreement between the prior and semantic data likelihood: even though the case C data is erroneous, it has a lower $\gamma_0$ than the correct data in case B, since the prior is higher in areas where case C's softmax likelihood is also large. 

\section{SIMULATION STUDY} \label{sec: section_sim}
SDA is studied in the context of a simulated planetary mineralogical survey task, building on the examples in Figs. \ref{fig:data_association_example} and \ref{fig:different_softmax_GMposterior}. 
Since the focus is on SDA algorithms, the simulations are based on template-based semantic interfaces for spatial object localization that were developed and validated with human-robot teams in previous collaborative sensing studies for similar 
problems \cite{nisarTRO, sample2012, lukeRAS}. The context for semantic sensing and simulation setup are first reviewed. The algorithms used to address SDA and metrics for evaluation are then described. The results are then presented and discussed.

\subsection{Futuristic Collaborative Planetary Exploration Scenario} \label{sec: section_2D}
We consider a Mars surface exploration mission, 
\cite{vasavada2014, mars2020overview}, 
where an autonomous ground robot is deployed to a region of the surface where the existence of scientifically interesting mineral deposits has been inferred {{\it a priori}} from overhead satellite data. The task is a follow up ground survey mission to acquire `ideal' specimens of different types and morphologies for detailed analysis \cite{tfong2010, aastha2020}. 
Due to the significant time delay and bandwidth limits for Earth-Mars communication, it is difficult to conduct deep space exploration efficiently with current practices that require major human involvement from combined Earth-based science and engineering teams to determine detailed robot action plans. 
%

These issues for planetary exploration can be addressed via telepresence, 
whereby astronauts on a space station orbiting the planet collaborate with remote explorers on the surface with minimal time delays \cite{lester2017telepresence}. 
In such an operation, 
exploration efficiency would be further enhanced if the astronauts aboard the space station were scientists \footnote{Historically speaking, Harrison Schmitt, the first scientist astronaut aboard Apollo 17, made numerous scientific discoveries on the Moon \cite{schmitt1973}}. As astronauts would not necessarily be robotics experts and are likely to be tasked with other duties during such as monitoring other surface and station operations, here we assume that they convey soft semantic data to the rover using domain-appropriate semantics. 
Since surface surveys can be modeled as Bayesian estimation and information gathering problems \cite{candela2017, arora2019}, this process would also allow exchange of key information grounded in discrete language-based descriptions of exploration environments and would support autonomous robotic reasoning. 

Suppose the human-robot team is required to locate 
four desired types of rock specimens (target objects):  
1) a large (above a certain diameter) calcite specimen; 2) a rounded (minimal number of edges/faces) calcite specimen; 3) a large pyroxene specimen; and 4) a rounded pyroxene specimen. 
The human sensor is a mission specialist astronaut with access to a camera-equipped aerial drone that travels a predetermined path to quickly scout the surface from a high altitude, while the autonomous ground rover searches for the desired specimens, 
with a limited-range visual object detector \cite{estlin2012, reena2017}. 

\begin{figure}[t]
    \centering
    \includegraphics[width=8.5cm]{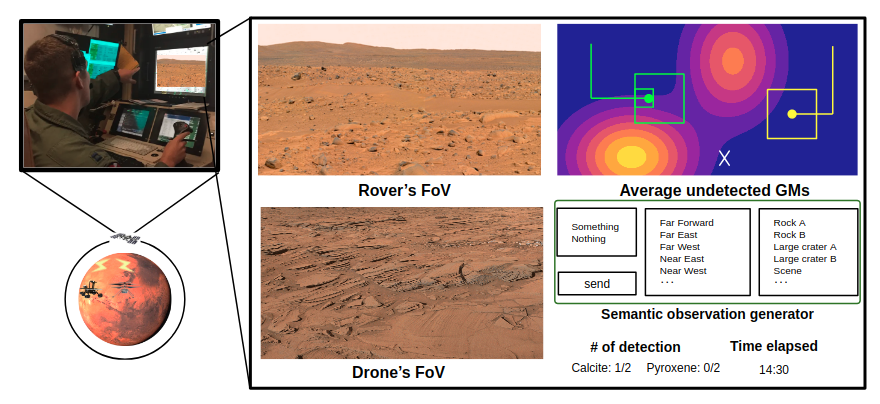}
    \caption{Notional UI for human-robot Martian surface survey: scientist astronaut (left) inspects mobile robot FOVs (middle) and provides semantic data by selecting combination of predefined semantic descriptor entries (right).}
    \label{fig:UI_of_mission}
    \vspace{-0.2in}
\end{figure}

Fig. \ref{fig:UI_of_mission} shows a {notional} UI for astronaut-robot communication. The rover and drone fields of view (FOVs) are displayed in the middle, while robot positions (green/yellow traces) 
and target specimens detected by the team (white X) are superimposed on the average pdf of undetected specimen locations (upper right). With this information at hand, the astronaut selects a combination of the predefined descriptor entries to create discrete semantic observations indicating the presence (or absence) of desirable specimens (lower right). 
 The rover fuses semantic soft data with onboard sensor data, and uses a simple open-loop path planning algorithm to navigate to the highest peak of the combined target state pdf \footnote{Though not the focus of this work, alternative POMDP-based closed-loop planning strategies \cite{lukeTRO,lukeRAS,lukeFUSION} could also be used to more formally exploit PSDA's GM-based continuous belief state representation and approximate optimal search strategies.}. 

Besides assisting with visual search, the astronaut still needs to work on other tasks as part of/in parallel with the exploration mission. Either out of habit or due to time pressure, the astronaut tends to provide observations of different mineral types and locations but omits information about the size/shape.
Thus, when monotonous Martian surface images are sent back for instance, there is a risk of \textcolor{black}{the astronaut providing} the 
false/wrongly-associated observations; if these are taken for granted by the rover, mission efficiency decreases. 
Unexpected accidents or delays (resulting from rover actions based on incorrect data) may also interrupt the mission. Therefore, the rover should carefully fuse human semantic data by considering possible sources of 
uncertainty. 

Formally summarizing in the terminology of Sec. \ref{sec: section_sda}, the SDA problem here concerns valid targets corresponding to ideal rock specimens with specific morphological features and mineral type. Each specimen of interest $i=1,\cdots,4$ to be found has an unknown location $x^i \in \mathbb{R}^2$ with corresponding prior $p(x^i)$. The ground robot uses a camera-based object detector that returns $\zeta_k=$ `Detection' or `No Detection' readings within a limited FOV at each time $k$. The human accesses this and another drone-based imaging system to use as the basis for generating semantic data $D_k$ 
according to a fixed descriptor set. When $D_k$ correctly describes a valid target specimen $i$, then $\theta_k=i$, otherwise $\theta_k=0$.

\subsection{Simulation Setup}
Based on the scenario described above,  simulations were performed on the $50[m]$-square search site shown in Fig. \ref{fig:sample_entire_map}. At each time step $k$, the ground rover receives visual object ``Detection (1)'' or ``No Detection (0)" data $\zeta_k$ data covering a $9 [m^2]$ area in front of it, taking the form of binary vector for each of the four different target mineral types described earlier (e.g. $\zeta_k = [0,0,0,1]^T$). The detection rate is high enough 
so that observation uncertainty and data association ambiguity for $\zeta_k$ can be ignored. For 
$p(\zeta_k|x)$, the 2D multi-modal softmax (MMS) model \cite{nisar2011} shown in Fig. \ref{fig:detector_likelihood} is used; the MMS model parameters are calculated offline \cite{nisar2018} and modified online based on the rover's position and orientation. VBIS and LWIS updates are respectively applied for rover ``Detection'' and ``No Detection" data. 
\begin{figure}[t]
    \centering
    \includegraphics[width=5.75cm]{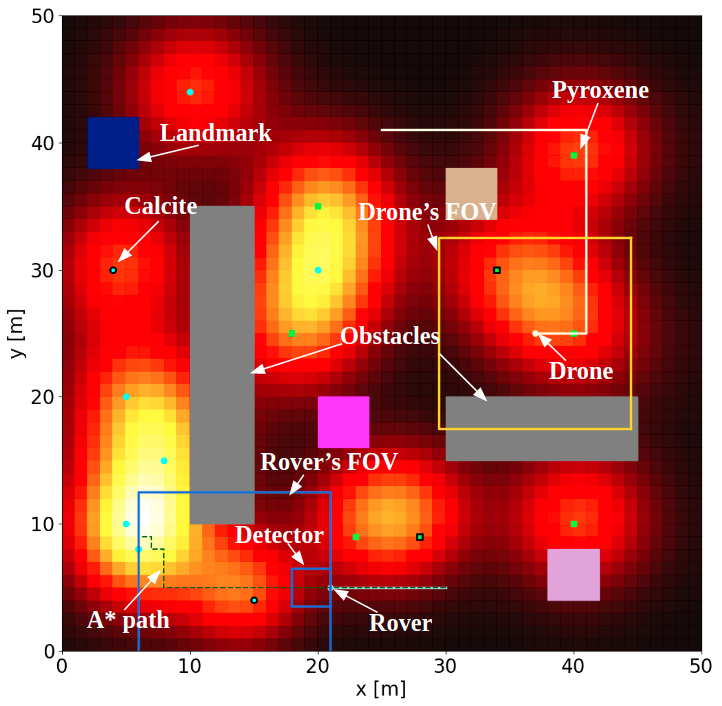}
    \caption{Average GMs of undetected targets overlaid on search site: rover moves toward highest pdf location, while drone travels pre-determined lawnmower path.}
    \label{fig:sample_entire_map}
    \vspace{-0.05in}
\end{figure}

\begin{figure}[t]
    \centering
    \includegraphics[width=8.6cm]{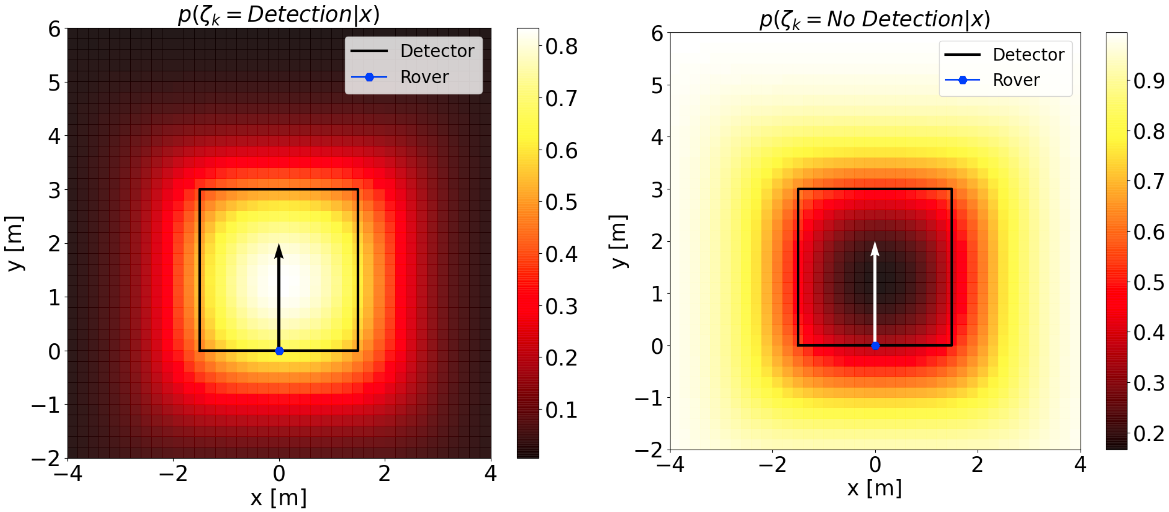}
    \caption{Rover detection (left) and non-detection (right) likelihoods.}
    \label{fig:detector_likelihood}
    \vspace{-0.15in}
\end{figure}

In addition, the autonomous rover has a navigation camera system 
that projects a $15 $ m$^2$ view in front of it {(representing a conservative vision field over images stitched from multiple forward cameras)}. 
Based on the image displayed in the Fig. \ref{fig:UI_of_mission} UI, a simulated astronaut occasionally provides either negative data (e.g. `No calcite is in the rover's FOV') or positive data (e.g. `Some good pyroxene is far left to the rover') for the rover. Such inputs occur every $8$ time steps with $80\%$ probability. Semantic data generated by the rover's navigation camera use the rover as a frame of reference (FOR). For negative semantic data provided by the astronaut using the rover's imaging system for reference, LWIS updates are applied relative to the ground rover's FOV without association amibiguity, since the human has zero probability of missed detections. 
For positive semantic data, the rover must address the SDA problem (with PSDA or another algorithm for comparison), due to human's non-zero false positive rate and incomplete target descriptions. 

Similarly, the simulated astronaut also provides semantic observations using the aerial drone's imager for reference. However, the drone is not used as a FOR, 
due to its fast rotation and constant vibration in the air. Instead, one of several environmental landmarks established previously within the search area are selected as a FOR. When no landmarks are in the drone's FOV or a detected object in the FOV is too far from a landmark, the astronaut simply provides `No good calcite/pyroxene is in the drone's FOV' and `Some good calcite/pyroxene is in the drone's FOV', respectively. 

These two types of structured soft semantic data are formed by selecting each element corresponding to four UI fields, $D_k = $ ``({\it Existence}) ({\it Object type}) is ({\it Direction}) ({\it FOR})", and observation likelihoods take values as shown in Fig. \ref{fig:different_softmax_GMposterior}. The false positive rate, i.e. $p(\theta_0)$, is higher when the astronaut uses the drone's imager (20\%) than the rover's imager (10\%) due to the drone's more unstable motion. 

The rover models the prior location of the four kinds of desired minerals using GM priors with $25$ components each. During the exploration, the rover computes the average GMs of undetected target objects 
{\allowdisplaybreaks
\abovedisplayskip = 2pt
\abovedisplayshortskip = 2pt
\belowdisplayskip = 2pt
\belowdisplayshortskip = 0pt
\begin{align} \label{eq: ave_maximum}
    p(x_k^{ave}) = \frac{1}{\textcolor{black}{N}} \sum_{q} p(x_k^{q} | \zeta_{1:k}, D_{rover, 1:k}, D_{drone, 1:k}),
\end{align}
}
\noindent where $D_{rover, 1:k}$ and $D_{drone, 1:k}$ are human semantic observations generated by reference to the rover's and drone's imagers, respectively. 
The maximum of (\ref{eq: ave_maximum}) is set as an intermediate goal point for an open-loop $A^*$ path planning algorithm (motion uncertainty is ignored along this path due to the rover's slow speed). The goal and the path of $A^*$ are recalculated if one of the undetected target objects is detected by the rover's detector, if the rover arrives at the goal, or if the astronaut provides semantic data. Similar to the approach used in Burks, et al. \cite{lukeTRO}, after fusing $\zeta_k$, $D_{rover, k}$ and $D_{drone, k}$, each target object posterior is compressed to $25$ mixands by first clustering mixtures into submixtures and then applying Runnalls' method \cite{runnalls2007}, which uses an upper bound on the KL divergence between pre-merge and post-merge submixture to select least dissimilar component pairs.

\subsection{SDA Approaches and Evaluation Metrics}
The following semantic data fusion modalities are compared: 1) Rover detector only (no human semantic data fused); 2) naive VBIS (No DA; trust all human semantic observations, i.e. the rover assumes that there are no false observations from human, and positive observations are mapped to both large and round types of minerals); 3) VBIS with naive DA; 4) Greedy-PSDA and 5) PSDA. For VBIS with naive DA, the simple DA method implemented in \cite{nisarTRO} is used, which treats all hypothesis association probabilities are equal and fixed when a positive semantic observation is provided. The naive DA posterior is as follows depending on whether a positive  (\ref{eq: positive_naive}) or negative (\ref{eq: negative_naive}) observation $D_{rover, k}/D_{drone, k}$ is received, 
{\allowdisplaybreaks
\abovedisplayskip = 2pt
\abovedisplayshortskip = 2pt
\belowdisplayskip = 2pt
\belowdisplayshortskip = 0pt
\begin{align} \label{eq: positive_naive}
    p_{naive}(x^i|D_{j\neq0}) &= \frac{1}{\textcolor{black}{N}} p(x^{i,*}) + \Big[1-\frac{1}{\textcolor{black}{N}}\Big] p(x^i), \\
    p_{naive}(x^i|D_{j = 0}) &= p(x^{i,*}).  \label{eq: negative_naive}
\end{align}
}
\noindent By comparing the coefficients of the first and second terms of (\ref{eq:final_psda_posterior}) and (\ref{eq: positive_naive}), we see that naive DA simplifies the calculation of association probabilities by applying simple heuristics, rather than a complete and rigorous Bayesian formulation {(though the latter can in fact be well-approximated for free as a side-product of VBIS calculations)}. 

For modalities $1$ through $5$, $20$ Monte Carlo simulation runs are executed with the different initial rover and drone configurations {to produce noisy observations that are sampled according to the respective likelihood models.} Since the data fusion method used in each modality is different, the posterior pdf of target objects is dissimilar, and as a result, the set of observations obtained during the search simulation is different as well as the intermediate goals to be achieved using $A^*$. All missions terminate when either the human-robot team does not find all four target objects (four small circles surrounded by a black frame in Fig. \ref{fig:sample_entire_map}) after $250$ time steps or when the autonomous rover cannot find a new goal position of $A^*$ path planning algorithm. This happens when pdfs of targets are concentrated so that $A^*$ start and goal points coincide. 

The simulation results are analyzed to assess several metrics. Firstly, we assess the extent to which the search mission described can be accomplished using only the ground rover's sensors alone, and establish the degree to which collaborative human-robot semantic sensing can truly benefit. We then assess whether the proposed PSDA method actually performs better than other data fusion modalities in terms of rate of mission success (i.e. the four target objects are found within the time limit), average numbers of detected objects, and average traveled distance by the ground rover when missions succeeds (similar to the metrics used in \cite{nisarTRO}, \cite{lukeRAS}). Then, we analyze the behavior of estimated posterior false alarm semantic data association probabilities {$\gamma_{PSDA, 0}$} to investigate how the rover reasons about the validity of human semantic observations in an online setting, when ground truth data for correct/incorrect semantic data are available. 

The true false positive  rate and assumed false positive (FP) rate for the semantic human sensor observations are initially assumed identical. However, in real human-robot interaction, there will be discrepancies between these values \cite{wang2012, nisar2015sketch}. Thus, we run additional simulations to assess the robustness of the PSDA method to potential mismatches between the true and algorithm-assumed FP values. Finally, we assess the sensitivity of PSDA to adverse placement of target objects near each other. 
This case should be considered because conventional PDA often struggles in situations where multiple targets have overlapping uncertainty bounds.

\subsection{Results: Difficulty of the Exploration Problem} \label{sec: difficulty_of_the_exploration}
We first establish the difficulty of the search task when only the ground robot's onboard detector is used to update the specimen distribution maps. As shown in Fig. \ref{fig:show_difficulty}, when only $\zeta_k$ observations  
are fused, the rover tends to miss objects along search paths, unless they are directly in front. This is because for almost the entire duration of the mission, only negative information is available for each target object and LWIS is performed, i.e. there is no significant difference between the prior and the posterior in a single observation and $A^*$ goal points are not updated drastically. Thus, it is difficult for the rover to find all targets by itself within the 
time limit.

\begin{figure}[t]
    \centering
    \includegraphics[width=5cm]{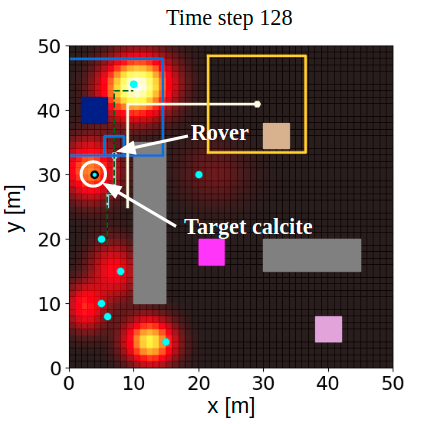}
    \caption{This snapshot shows the difficulty of the multi-object search mission: autonomous rover passes by and misses calcite target due to small onboard detector area. Coarse but wide-range human semantic sensing provides complementary data for fast and efficient exploration.}
    \label{fig:show_difficulty}
    \vspace{-0.15in}
\end{figure}

\begin{figure*}[htbp]
        \centering
        \includegraphics[width=17cm]{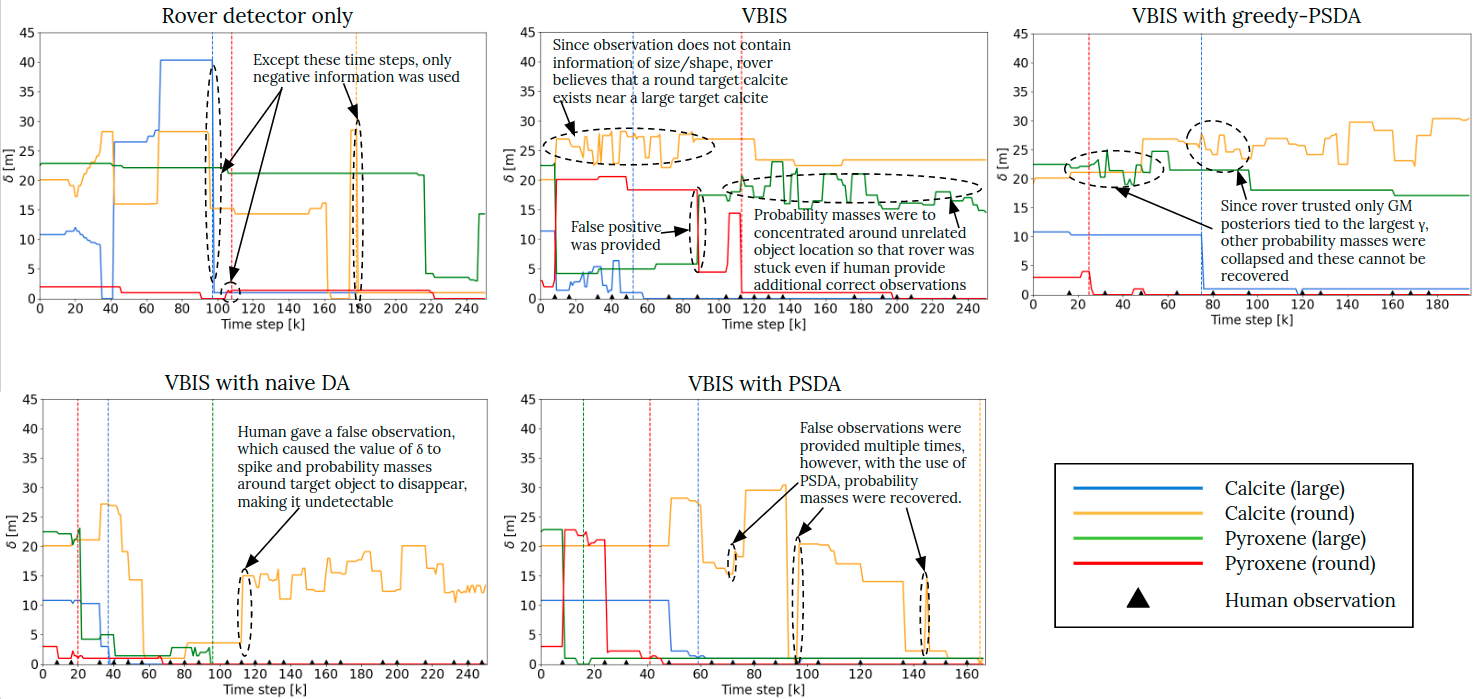}
        \caption{Typical two-norm MAP-estimated position errors of target objects \textcolor{black}{$\delta_k = ||x_{true}-\hat{x}_k||^2$, where $\hat{x}_k = \arg\max p(x|\zeta_{1:k}, D_{1:k})$} over time steps: Black triangles on the axes represent when the simulated astronaut provides semantic observation $D_{rover}$ and $D_{drone}$. Vertical dashed lines indicate when each target object is detected by the rover detector. \textcolor{black}{In contrast to the baselines, the results using the proposed PSDA algorithm (bottom center) successfully avoided biasing/collapsing probability masses when incorrect semantic observations were provided.} }
        \label{fig:delta_comparison}
\end{figure*}

\subsection{Results: Performance in Different Fusion Modalities} 
The results of 20 Monte Carlo runs for all data fusion modalities are summarized in Table \ref{tab:compare_different_fusion}. Also, the time history of the maximum a posteriori (MAP) estimated target object position error $\delta_k = ||x_{true} - \hat{x}_k||_2$, where $\hat{x}_k = \arg \max \ p(x|\zeta_{1:k}, D_{1:k})$ are shown in Fig. \ref{fig:delta_comparison} for typical search runs, to show how information accumulates in the GM posterior pdfs for each target object over time. Although fusion modality 1 (rover detector only) succeeds in finding several targets objects, it could not find all within the time limit in all Monte Carlo runs, as described in Sec. \ref{sec: difficulty_of_the_exploration}. When no data association is used and the human is treated as an oracle (modality 2), none of the Monte Carlo runs results in a successful mission. Additionally, the average number of target objects found is lower than in modality 1. 
This follows from naive processing of incorrect soft data, which cause probability mass to move away from actual target locations.

As for VBIS with naive DA (modality 3), when positive semantic observations are provided with multiple possible targets, the updates via (\ref{eq: positive_naive}) are gentler, so that probability around proper regions recovers even when incorrect soft data is provided. However, when $\textcolor{black}{N} = 1$, since naive DA does not account for erroneous semantic observations, the rover often got stuck in wrong areas, as shown in Fig. \ref{fig:failure_naive_PDA}, and only $14$ out of $20$ Monte Carlo runs succeed. 
\begin{table}[t]
\centering
\caption{Comparison of different data fusion modalities}
\label{tab:compare_different_fusion}
\begin{tabular}{|c|c|c|c|}
\hline
\multicolumn{1}{|l|}{} & \begin{tabular}[c]{@{}c@{}} \# of successess \\ out of 20 runs\end{tabular} & \begin{tabular}[c]{@{}c@{}}Average \# of \\ detected objects\end{tabular} & \begin{tabular}[c]{@{}c@{}}Average traveled \\ distance when \\ succeeded (m)\end{tabular} \\ \hline
Only detector    & 0 & 2.4 & - \\ \hline
No DA            & 0 & 1.4 & - \\ \hline
Naive DA         & 14& 3.4 & 148.21 \\ \hline
Greedy-PSDA      & 5& 2.15& 113.40 \\ \hline
PSDA             & 19& 3.95& 143.26 \\ \hline
\end{tabular}
\vspace{-0.2in}
\end{table}
When GM posteriors are formed via greedy-PSDA, the rover is able to move towards targets more quickly, since only one hypothesis from the GM posterior is used and all other hypotheses are excluded. In fact, the average traveled distance when the missions succeeded using greedy-PSDA ($113.40$ m) was shorter than that of other methods. However, when the Mahalanobis distance between the mean of GM posterior associated with the largest {$\gamma_{PSDA}$} and a target is large enough, the probability mass around the target disappeared and missions terminated as in modality 2. Therefore, the mission success rate was much lower than that of naive DA and PSDA methods. Finally, when PSDA was used, although the human provides erroneous data, (Fig. \ref{prob_mass_recovery_PDA}, left), the rover succeeded more often in recovering probability around actual target locations (Fig. \ref{prob_mass_recovery_PDA}, center and right). Thanks to this ability, all targets were found in $19$ out of $20$ Monte Carlo runs and improved on naive DA's travel distance. 
\begin{figure}[t]
    \centering
    \includegraphics[height=5cm]{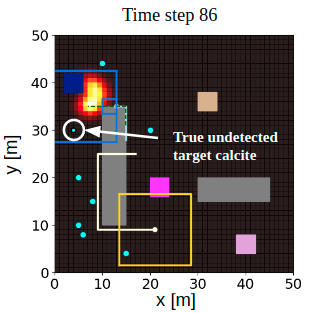}
    \caption{Probability mass depletion with naive DA: after the simulated astronaut provided `Calcite is near ahead of you', the rover trusted the observation and wrongly updated the target pdf. Even after observing 'No detection' at the region, probability mass around actual target did not recover.}
    \label{fig:failure_naive_PDA}
    \vspace{-0.2in}
\end{figure}

\begin{figure*}[t]
 \begin{center}
  \includegraphics[height=5.0cm]{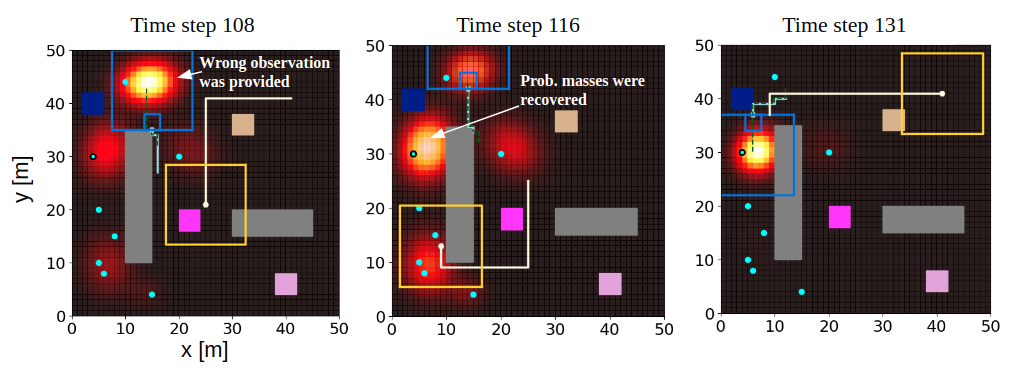}
  \caption{Probability mass recovery with PSDA: astronaut provides `Calcite is near ahead of you' and rover believed the human provided wrong information with probability $\gamma_{PSDA, 0}=0.47$. Thus, \textcolor{black}{contrary to Fig. \ref{fig:failure_naive_PDA},} some initial probability mass remained around the true target location (left); once the rover reached the wrong \textcolor{black}{estimated location}, the probability around the true calcite location fully recovered (center) and the target was successfully detected (right).}
  \label{prob_mass_recovery_PDA}
 \end{center}
 \vspace{-0.25in}
\end{figure*}

\subsection{Results: Analysis of {$\gamma_{PSDA, 0}$}}

In PSDA, recall that {$\gamma_{PSDA, 0}$} given by (\ref{eq: gamma_0i}) is the probability that a semantic datum does not describe to any object of interest. \textcolor{black}{Since $\sum_{i=0}^{N}\gamma_{PSDA,i}=1$ for each $D_j$}, the value of {$\gamma_{PSDA, 0}$} \textcolor{black}{varies with time}; in particular, {$\gamma_{PSDA, 0}$} is high (low) when $\sum_{s=1}^{\textcolor{black}{N}} p(D_j, \theta_s)$ is small (large). Examining  (\ref{eq:joint_D_theta_i}), when $p(D_j|\theta_s, x^s)$ and $p(x^s)$ do not overlap enough, $p(D_j, \theta_s)$ does not have a large value, resulting in a large {$\gamma_{PSDA, 0}$}. 
Therefore, in our study, we can say that PSDA is more {\it conservative} when the majority of GM priors is far from the peak of softmax likelihood and {\it slightly optimistic} when the majority of GM priors is close enough to the peak of softmax likelihood. An illustrative example is shown in Fig. \ref{fig:gamma_comparison}. Although the simulated astronaut provides erroneous data, this is correctly inferred by PSDA: at time $72$, $D_{rover, 72}$ is `Some good calcite is in front of you' resulting in $\gamma_{PSDA, 0}=0.123$, while at time step $96$, $D_{rover, 96}$ is `Some good calcite is near right to you' with $\gamma_{PSDA, 0}=0.413$.
\begin{figure}[t]
    \centering
    \includegraphics[height=4.5cm]{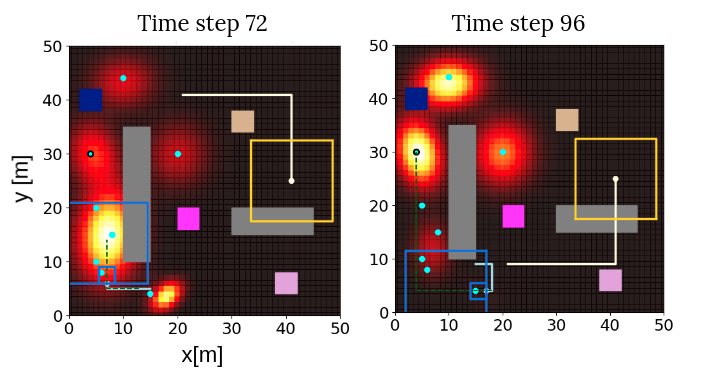}
    \caption{Scenario across time steps for comparison of the values of {$\gamma_{PSDA, 0}$}}
    \label{fig:gamma_comparison}
    \vspace{-0.1in}
\end{figure} 

\subsection{Results: Robustness to True vs. Assumed FP Rate}
To check the robustness of PSDA against deviations between true and assumed false positive rates (FP), 20 extra Monte Carlo simulation runs were executed with different combinations of true and assumed FP. As seen in Fig. \ref{fig:different_FP_combination}, when the rover assumes optimistic FP values (magenta boxes), the mission success rate of finding all four target rock specimens is low. On the contrary, when the rover is conservative or has perfect knowledge of FP, the number of mission successes is higher, except when both true and assumed FP is $0.5$.  This is mainly because the simulated human provides wrong observations too often and the autonomous rover needs more time to find all target objects. This practically suggests 
being conservative about human semantic observations FP rates. 

\begin{figure}[htbp]
    \centering
    \includegraphics[height=4.3cm]{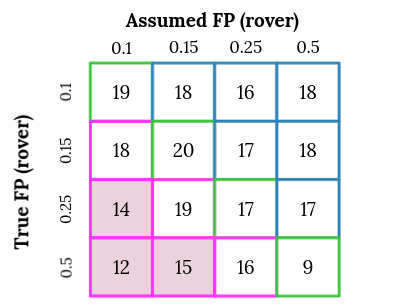}
    \caption{Effect of assumed/true FP rates on number of search success (out of 20 Monte Carlo runs). There are three categories where the rover's knowledge of FP rate is: perfect (green); conservative (blue); and optimistic (magenta).}
    \label{fig:different_FP_combination}
    \vspace{-0.15in}
\end{figure}

\subsection{Results: Robustness for Adverse Object Placement}

Until now, target locations were identical across all simulation runs.  
Fig. \ref{fig:challenging_config} shows a more challenging configuration for object locations, where inconsistent initial GM priors assign low probability to true locations. This scenario is especially hard because most objects (including non-targets) are close to each other, increasing the possibility that the simulated astronaut provides erroneous data. As the rover approaches an unrelated object due to incorrect semantic data, 
it erases the surrounding probability masses, so that the new $A^*$ goal tends to appear further away from the true targets each time following subsequent fusion update, causing the robot to travel further than necessary. Even in this case, all 4 objects were found within the time limit in 14/20 runs using PSDA. 

\begin{figure}[t]
    \centering
    \includegraphics[height=4.3cm]{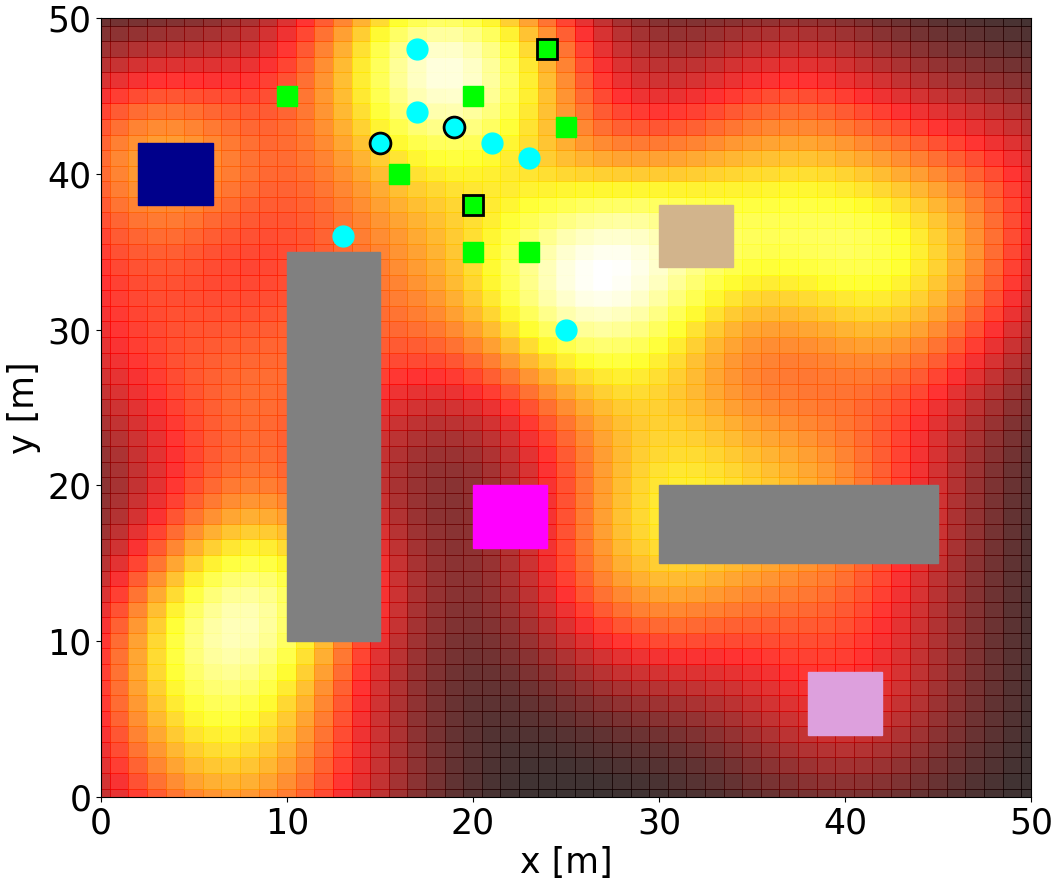}
    \caption{Challenging search configuration with non-consistent GM priors.}
    \label{fig:challenging_config}
\end{figure}

\section{CONCLUSIONS} \label{sec: section_conc}
This paper presented semantic data association (SDA) as the class of state estimation problems where the issues of hybrid data fusion and data association problems overlap, and derived a novel probabilistic semantic data association (PSDA) algorithm to deal with possibly erroneous and ambiguous semantic human observations. PSDA can exploit the byproducts of semantic data fusion algorithms that enable approximate updating of prior pdfs with semantic observation likelihoods. This feature was leveraged to generate association hypothesis probabilities for single Gaussian and Gaussian mixture state priors when softmax likelihood functions are used
through the VBIS approximation devised in \cite{nisarTRO}. PSDA was integrated into a recursive Bayesian inference problem for a simulated rover-based multi-object search and shown to outperform other SDA algorithms on several metrics. The results also showed that resulting association hypotheses probabilities tend to be {\it conservative} or {\it slightly optimistic}, according to the extent of overlap between priors and likelihoods. PSDA was verified to be robust to the mismatch between true and assumed false positive rates as well as to adverse priors and state configurations. 
The grounding of these results within a rigorous probabilistic framework for SDA provides a useful baseline for further assessment and comparison of PSDA and other possible SDA algorithms in different real-world collaborative human-robot sensing problems.

\section*{Acknowledgments}
S. Wakayama was supported by the Masason Foundation.

{\appendix[Derivation of the multi-state PSDA posterior ] \label{appx: derivation}
In (\ref{eq:joint_D_theta_multi}), since the value of $\theta$ is given in the first term, only the pdf whose index is $i$ is updated by $D_j$ when $i \neq 0$ {(i.e. when $D_j$ actually refers to a state of interest)}. From the independence of priors of $x^1, \cdots, x^{\textcolor{black}{N}}$, 
{\allowdisplaybreaks
\abovedisplayskip = 2pt
\abovedisplayshortskip = 2pt
\belowdisplayskip = 2pt
\belowdisplayshortskip = 0pt
\begin{align} \label{eq:joint_posterior_given_theta}
    p(X|D_j, \theta_i) = p(x^i|D_j, \theta_i) \prod_{l \neq i} p(x^l).
\end{align}
}
{On the contrary}, when $i = 0$ {(i.e. when $D_j$ does not actually refer to a state of interest)}, $p(X|D_j, \theta_0)$ is simply reduced to 
{\allowdisplaybreaks
\abovedisplayskip = 2pt
\abovedisplayshortskip = 2pt
\belowdisplayskip = 2pt
\belowdisplayshortskip = 0pt
\begin{align} \label{eq:joint_posterior_given_theta0}
    p(X|D_j, \theta_0) = p(X) = \prod_{l=1}^{\textcolor{black}{N}} p(x^l).
\end{align}
}
Regarding the second term of (\ref{eq:joint_D_theta_multi}), the joint distribution $p(D_j, \theta_i)$ is the marginalization of another joint distribution $p(D_j, \theta_i, X)$, that is, 
{\allowdisplaybreaks
\abovedisplayskip = 2pt
\abovedisplayshortskip = 2pt
\belowdisplayskip = 2pt
\belowdisplayshortskip = 0pt
\begin{align}
    p(D_j, \theta_i) = \int_{x^1} \cdots \int_{x^{\textcolor{black}{N}}}p(D_j, \theta_i, X)dx^1\cdots dx^{\textcolor{black}{N}},
\end{align}
}
\noindent where $p(D_j, \theta_i, X) = p(D_j|\theta_i, X)p(\theta_i, X)$. When $i \neq 0$, 
$p(D_j|\theta_i, X)$ is reduced to $p(D_j|\theta_i, x^i)$. Since $\theta_i, x^1, \cdots, x^{\textcolor{black}{N}}$ are all independent before receiving $D_j$, $p(D_j, \theta_i)$ is rewritten as 
{\allowdisplaybreaks
\abovedisplayskip = 2pt
\abovedisplayshortskip = 2pt
\belowdisplayskip = 2pt
\belowdisplayshortskip = 0pt
\begin{align}
    p(D_j, \theta_i) = \int_{x^i}p(D_j|\theta_i, x^i)p(x^i)p(\theta_i)dx^i, \ i \neq 0. \label{eq:joint_D_theta_i}
\end{align}
}
On the other hand, when $i = 0$, 
$p(D_j, \theta_0)$ is transformed to
{
\allowdisplaybreaks
\abovedisplayskip = 2pt
\abovedisplayshortskip = 2pt
\belowdisplayskip = 2pt
\belowdisplayshortskip = 0pt
\begin{align} 
    p(D_j, \theta_0) &= \int_X p(D_j,\theta_0, X) dX \nonumber \\
    &= \int_X p(D_j|\theta_0, X)p(\theta_0, X)dX \nonumber \\
    &= \int_X \frac{p(D_j, \theta_0, X)}{p(\theta_0, X)}p(\theta_0)p(X)dX \nonumber \\
    &= \int_X \frac{p(X|D_j,\theta_0)p(D_j,\theta_0)}{p(\theta_0)p(X)}p(\theta_0)p(X)dX \nonumber \\
    &= \int_X \frac{p(D_j, \theta_0)}{p(\theta_0)} p(\theta_0)p(X)dX \nonumber \\
    &= \frac{p(D_j, \theta_0)}{\sum_D p(D, \theta_0)} p(\theta_0) = \frac{1}{H}p(\theta_0), \label{eq:d_j_theta_0_multi}
\end{align}
}
where $p(\theta_0)$ is the false report rate. Substituting (\ref{eq:joint_posterior_given_theta}), (\ref{eq:joint_posterior_given_theta0}), (\ref{eq:joint_D_theta_i}), and (\ref{eq:d_j_theta_0_multi}) into (\ref{eq:joint_D_theta_multi}) yields the joint target posterior (\ref{eq: joint_posterior_general}).
}


 

\bibliographystyle{IEEEtran}
\bibliography{root.bib}

\begin{IEEEbiography}
[{\includegraphics[width=1in,height=1.25in,clip,keepaspectratio]{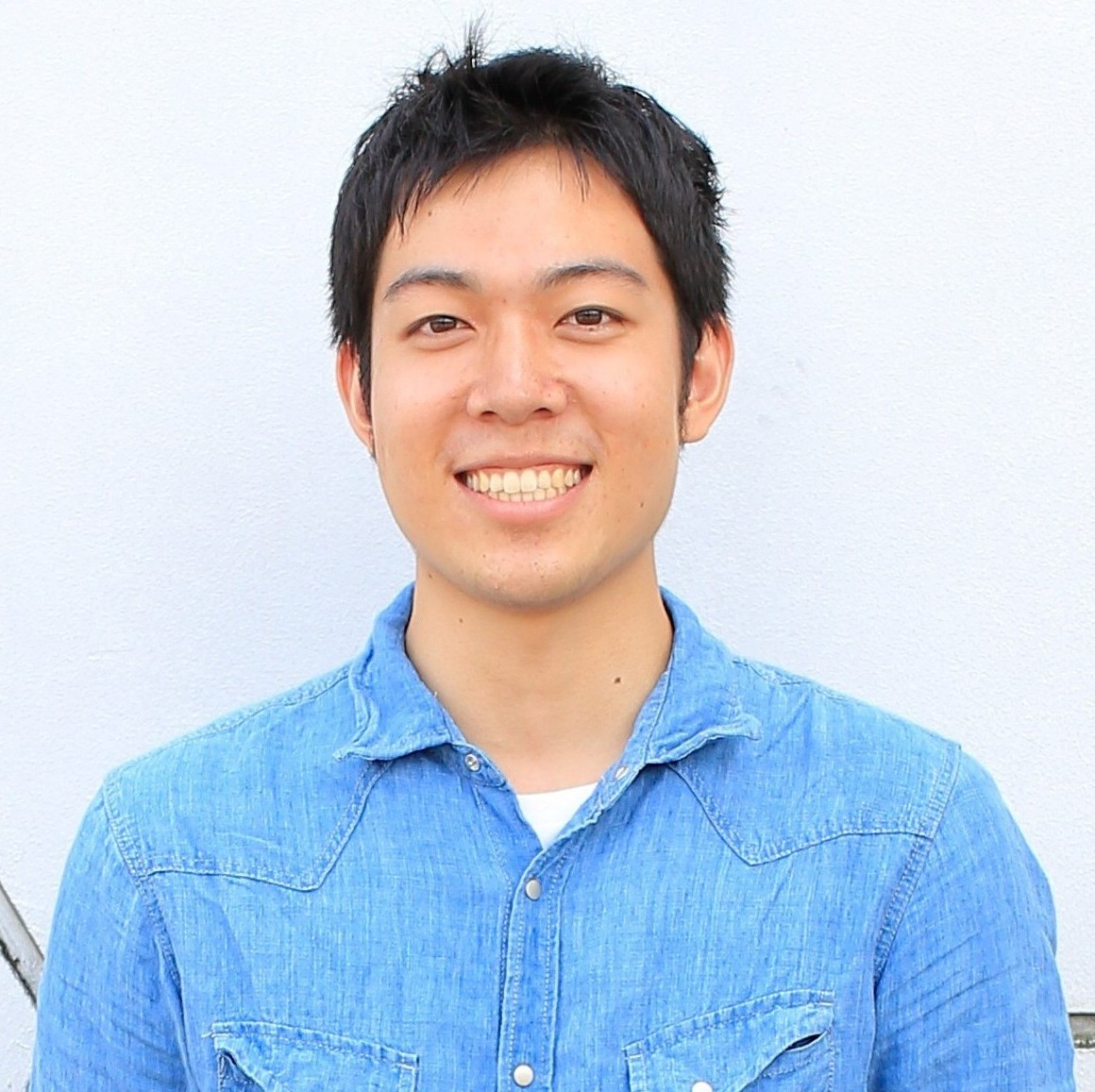}}]{Shohei Wakayama}
is a Ph.D. student in the Smead Aerospace Engineering Sciences Department at the University of Colorado Boulder. He received his B.S. in Mechanical Engineering from Kyushu University in 2018. His research interests lie in decision making under uncertainty, state estimation, and human-robot interaction for robotic exploration of unknown remote environments.  
\end{IEEEbiography}

\begin{IEEEbiography}
[{\includegraphics[width=1in,height=1.25in,clip,keepaspectratio]{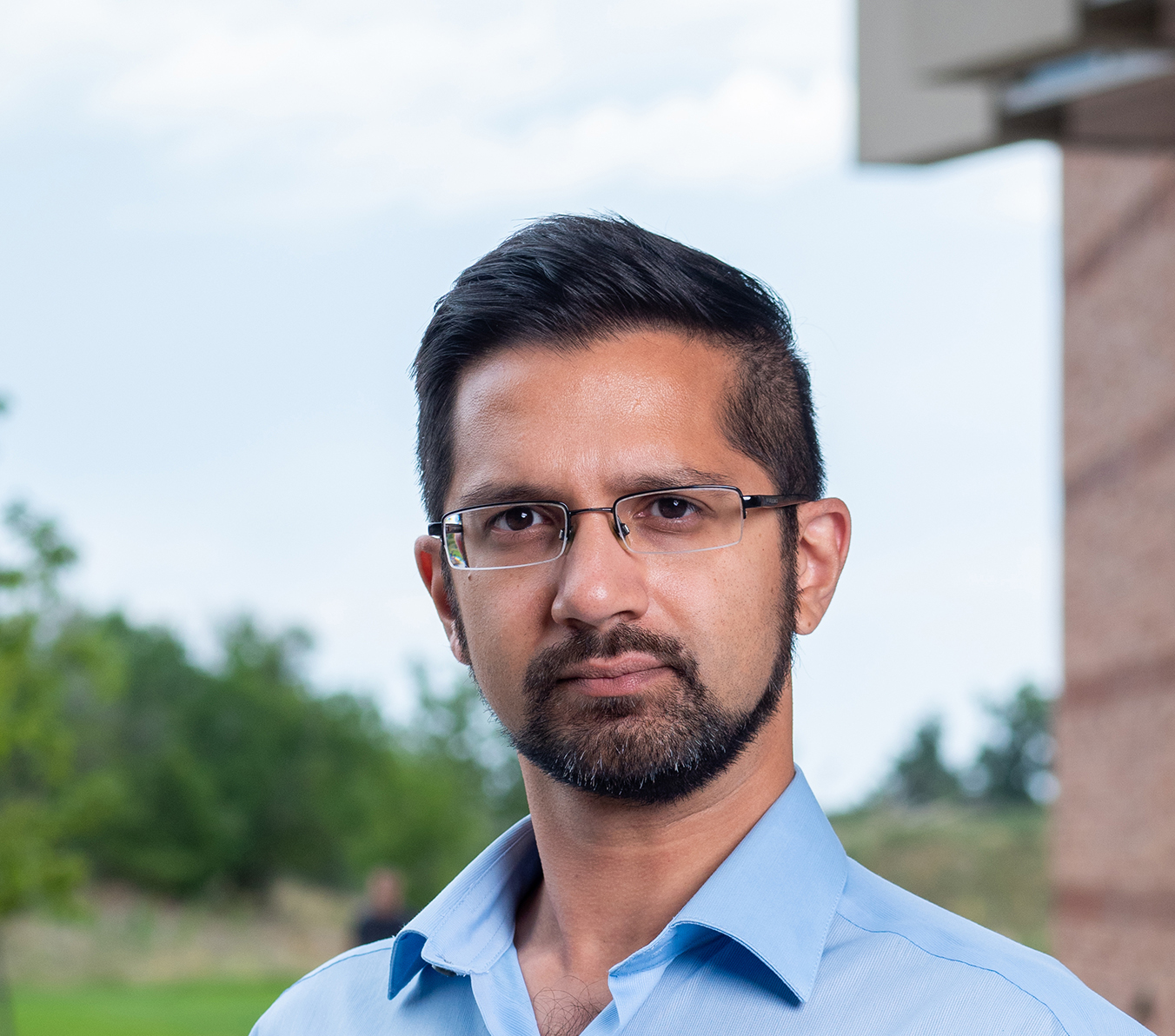}}]{Nisar Ahmed}
is an Associate Professor and H.J. Smead Faculty Fellow in the Smead Aerospace Engineering Sciences Department at the University of Colorado Boulder. 
He 
directs the Cooperative Human-Robot Intelligence (COHRINT) Lab, which researches probabilistic modeling, estimation and control of autonomous systems, human-robot/machine interaction, sensor fusion, and decision-making under uncertainty. 
\end{IEEEbiography}

\vspace{11pt}

\vfill

\end{document}